\documentclass[moor,nonblindrev]{informs3opt} 

\usepackage{endnotes}
\let\footnote=\endnote

\usepackage{natbib}
\NatBibNumeric
\def\bibfont{\small}%
\bibpunct[, ]{[}{]}{,}{n}{}{,}%

\usepackage{natbib}
\usepackage{titlesec,soul}
\usepackage[titletoc,toc,title]{appendix}
\usepackage{graphicx}
\usepackage{subfigure}
\usepackage{multirow}

\usepackage[hidelinks,bookmarksnumbered=true]{hyperref}
\usepackage{todonotes,bm}
\usepackage{tikz}
\usepackage{url}
\usepackage{amssymb}
\usepackage{psfrag}
\usepackage{amsmath}
\usepackage{setspace}
\newcommand{\exclude}[1]{}
\usepackage{bm}
\usepackage[flushleft]{threeparttable}
\usepackage{algorithm}
\usepackage{algpseudocode}
\algdef{SE}[DOWHILE]{Do}{doWhile}{\algorithmicdo}[1]{\algorithmicwhile\ #1}%
\algnewcommand{\Or}{\textbf{or}}
\algnewcommand{\And}{\textbf{and}}

\usepackage{thmtools} 
\usepackage{thm-restate}

\usepackage{cleveref}

\declaretheorem[name=Theorem]{theorem}
\declaretheorem[name=Proposition]{proposition}

\declaretheorem[name=Claim]{claim}
\declaretheorem[name=Corollary]{corollary}

\declaretheorem[name=Definition]{definition}
\declaretheorem[name=Example]{example}

\def\Re{\mathbb{R}}

\def\hat{\widehat}

\def\Re{{\mathbb R}}

\newcommand{\e}{\mathbf{e}}

\DeclareMathOperator{\supp}{supp}
\DeclareMathOperator{\diag}{diag}
\DeclareMathOperator{\Diag}{Diag}

\DeclareMathOperator{\tr}{tr}

\renewcommand{\S}{\mathcal{\bm S}}
\renewcommand*{\qed}{\hfill\ensuremath{\square}}
\newcommand*{\qedA}{\hfill\ensuremath{\diamond}}

\usepackage{enumerate,comment}

\allowdisplaybreaks

\begin{document}
	\TITLE{Exact and Approximation Algorithms for Sparse PCA}
	
	\ARTICLEAUTHORS{%
		\AUTHOR{Yongchun Li}
		\AFF{Department of Industrial \& Systems Engineering, Virginia Tech, Blacksburg, VA 24061, \EMAIL{liyc@vt.edu} \URL{}}
		\AUTHOR{Weijun Xie}
		\AFF{Department of Industrial \& Systems Engineering, Virginia Tech, Blacksburg, VA 24061, \EMAIL{wxie@vt.edu} \URL{}}
	} 
	\RUNAUTHOR{Yongchun Li and Weijun Xie}
	
	\RUNTITLE{Exact and Approximation Algorithms for Sparse PCA}
	\ABSTRACT{%
		Sparse PCA (SPCA) is a fundamental model in machine learning and data analytics, which has witnessed a variety of application areas such as finance, manufacturing, biology, healthcare. To select a prespecified-size principal submatrix from a covariance matrix to maximize its largest eigenvalue for the better interpretability purpose, SPCA advances the conventional PCA with both feature selection and dimensionality reduction. Existing approaches often approximate SPCA as a semi-definite program (SDP) without strictly enforcing the important cardinality constraint that restricts the number of selected features to be a constant. To fill this gap, we propose two exact mixed-integer SDPs (MISDPs) by exploiting the spectral decomposition of the covariance matrix and the properties of the largest eigenvalues. We then analyze the theoretical optimality gaps of their continuous relaxation values and prove that they are stronger than that of the state-of-art one. We further show that the continuous relaxations of two MISDPs can be recast as saddle point problems without involving semi-definite cones, and thus can be effectively solved by first-order methods such as the subgradient method. Since off-the-shelf solvers, in general, have difficulty in solving MISDPs, we approximate SPCA with arbitrary accuracy by a mixed-integer linear program (MILP) of a similar size as MISDPs. The continuous relaxation values of two MISDPs can be leveraged to reduce the size of the proposed MILP further. To be more scalable, we also analyze greedy and local search algorithms, prove their first-known approximation ratios, and show that the approximation ratios are tight. Our numerical study demonstrates that the continuous relaxation values of the proposed MISDPs are quite close to optimality, the proposed MILP model can solve small and medium-size instances to optimality, and the approximation algorithms work very well for all the instances. Finally, we extend the analyses to Rank-one Sparse SVD (R1-SSVD) with non-symmetric matrices and Sparse Fair PCA (SFPCA) when there are multiple covariance matrices, each corresponding to a protected group.
	}%
	
	
	\KEYWORDS{Sparse PCA, Largest Eigenvalue, Mixed-Integer Program, Semi-definite Program, Greedy, Local Search, SVD, Fairness}
	
	\maketitle
	
	\section{Introduction}
	This paper studies the sparse principal component analysis (SPCA) problem of the form
	\begin{align}\label{spca}
	\text{\rm (SPCA)} \quad w^{*} := \max_{\bm{x} \in \Re^n} \left \{\bm{x}^{\top}\bm{A}\bm{x}: {||\bm{x}||_{2}=1}, {||\bm{x}||_0 = k} \right \},
	\end{align}
	where 
	the symmetric positive semi-definite matrix $\bm{A} \in \Re^{n\times n}$ denotes the sample covariance out of a dataset with $n$ features and the integer $k \in [n]$ denotes the sparsity of its first principal component (PC). 
	In SPCA \eqref{spca}, the objective is to select the best size-$k$ principal submatrix from a covariance matrix $\bm{A}$ with the maximum largest eigenvalue.
	Compared to the conventional PCA, the extra zero-norm constraint $||\bm{x}||_0 = k$ in SPCA \eqref{spca} restricts the number of features of the first PC $\bm x$ to be $k$ most important ones. In this way, SPCA improves the interpretability of the obtained PC, which has been shown as early as \citet{jeffers1967two} in 1967. It is also recognized that SPCA can be more reliable for large-scale datasets than PCA, where the number of features is far more than that of observations \citep{zhang2011large}. These advantages of SPCA have benefited many application fields such as biology, finance, cloud computing, and healthcare, which frequently deal with datasets with a massive number of features (see, e.g., \citep{chaib2015informative, jiang2012family,luss2010clustering,naikal2011informative}). 
	
	
	\subsection{Relevant Literature}
	Our paper contributes to relevant literature on SPCA from three aspects: exact mixed-integer programs, convex relaxations, and approximation algorithms.
	
\noindent	\textbf{Exact Mixed-Integer Programs:}
	As shown in formulation \eqref{spca}, SPCA is highly non-convex- maximizing a convex function subject to two nonconvex constraints (i.e., an $L_2$ equality constraint and an $L_0$ equality constraint). Albeit superior to traditional PCA, SPCA \eqref{spca} is notoriously known to be computationally expensive; see, e.g., the complexity analysis and inapproximability results in \citet{magdon2017np}.
	As a result, the equivalent formulations and algorithms for exactly solving SPCA are quite limited in the literature (see, e.g., \citep{berk2019certifiably, gally2016computing,moghaddam2006spectral}). \citet{moghaddam2006spectral} introduced a branch and bound method to solve SPCA, and they pruned redundant nodes using the eigenvalue of principal submatrices and a greedy algorithm. Recently, \citet{berk2019certifiably} embedded various upper and lower bounds into this branch and bound framework, which could efficiently prune nodes and quickly certificate the optimality for quite a few instances. It is worthy of mentioning that \citet{gally2016computing} proposed a MISDP (MISDP) formulation for SPCA. Our second MISDP formulation differs from \citet{gally2016computing} by deriving two strong conic valid inequalities. Another interesting work can be found in \citet{dey2018convex}, where the authors developed approximate convex integer programs for SPCA with an optimality gap of $(1+\sqrt{k/(k+1)})^2$. 
	Quite differently, we propose two exact MISDP formulations and one approximate mixed-integer linear program (MILP) for SPCA from novel perspectives of analyzing the largest eigenvalue. 
	Specifically, the proposed MILP formulation can be arbitrarily close to the optimal value of SPCA, and it can be directly solved by off-the-shelf solvers such as Gurobi. 
	
	\textbf{Convex Relaxations:} Besides solving exact SPCA, researchers have also actively sought to explore effective convex relaxations. A common approach in literature is to develop SDP relaxations for SPCA (see e.g., \citep{amini2008high,d2005direct,d2008optimal,d2012approximation,zhang2012sparse}). 
	 Albeit convex, solvers often have difficulty in solving large-scale instances of SDP formulations (e.g., $n=\Omega(100)$). 
	The computational challenge of these SDP problems urgently calls for more effective methods to compute the relaxation values for SPCA. From a different angle, this paper solves the continuous relaxations of the proposed MISDP formulations as the maximin saddle point problem, where the subgradient method enjoys a $O(1/{T})$ rate of convergence \cite{nedic2009subgradient} based on Euclidean projections. Surprisingly, we further show that the projection oracle of the subgradient method is a second-order conic program rather than an SDP and thus can be easily dealt with. 
	
\noindent	\textbf{Approximation Algorithm:} Another early thread of research on SPCA is the development of high-quality heuristics for solving SPCA to near optimality such as greedy algorithm \citep{d2008optimal,he2011algorithm}, truncation algorithm \cite{chan2016approximability}, power method \citep{journee2010generalized}, and variable neighborhood search method \citep{carrizosa2014rs}. In particular, the truncation algorithm in \cite{chan2016approximability} so far provides the best-known approximation ratio $O(n^{-1/3})$, which can be easily implemented to generate a feasible solution for SPCA. This paper investigates the greedy and local search algorithms and proves their first-known approximation ratios $O(1/k)$ for SPCA. 
	
	\subsection{Summary of Contributions}
	We observe that when the support of $\bm x$ has been successfully identified, SPCA \eqref{spca} reduces to the conventional PCA finding the largest eigenvalue and eigenvector of a size-$k$ principal submatrix of $\bm A$. This fact motivates us to derive two equivalent MISDP formulations and an approximate MILP of SPCA. Below is a summary of the main contributions in this paper.
	\begin{enumerate}[(i)]
	\item For each formulation, we derive the theoretical optimality gap between its continuous relaxation value and the optimal value of SPCA. 
		\item Our first MISDP formulation inspires us to derive closed-form expressions of the coefficients of valid inequalities, which can be efficiently embedded into the branch and cut algorithms; 
		\item We show that the subgradient method can be adapted to ease the computational burden of obtaining MISDP continuous relaxation values with $O(1/{T})$ rate of convergence. These continuous relaxations values can further help reduce the size of MILP;
		\item The continuous relaxation of our second MISDP formulation is proven to be stronger than the one proposed in \citet{d2005direct};
\item The proposed MILP formulation has a similar size as two MISDPs and can be directly solved using many existing solvers;
		\item We prove and demonstrate the tightness of the first-known approximation ratios for the greedy and local search algorithms;
		\item Our analyses can be extended to the Rank-one Sparse SVD (R1-SSVD), which aims to compute the largest singular value of the possibly non-symmetric matrix $\bm A$ with the sparsity constraints on its left-singular and right-singular vectors separately; and
		\item We extend the second MISDP formulation to Sparse Fair PCA (SFPCA), where the covariance matrices are observed from multiple protected groups.
	\end{enumerate}
	Our contributions have both theoretical and practical relevance. Theoretically, we contribute three exact mixed-integer convex programs to SPCA. Practically, our MILP formulation can either attain optimal solutions for SPCA, improve the continuous relaxations, or find better-quality feasible solutions for small and medium-size instances. We apply the computationally efficient subgradient method to solving the continuous relaxations of the proposed MISDPs, as well as deriving their theoretical optimality gaps.
	We also develop two scalable approximation algorithms to solve SPCA to near optimality and prove their approximation ratios. Our proposed algorithms have been demonstrated to be successfully applied to large-scale data analytics problems, such as identifying key features for the drug abuse problem.
	We further extend the analyses to R1-SSVD and SFPCA. All the theoretical contributions are summarized in \Cref{table:spca}. 
	\begin{table}[htbp]
				\caption{Summary of Theoretical Contributions} 
		\centering
		\label{table:spca}
		\begin{threeparttable}
			\setlength{\tabcolsep}{3pt}\renewcommand{\arraystretch}{1.5}\small
			\begin{tabular}{ |c | c |c| }
				\hline
				\multicolumn{1}{|c|}{\textbf{Problem}}	
				&\multicolumn{1}{c|}{\textbf{Exact Mixed Integer Program}} & \multicolumn{1}{c|}{\textbf{Optimality Gap\tnote{2}}} \\ 
				\hline
				\multirow{3}{4.5em}{SPCA}&	MISDP \eqref{sdp_two_eq} & $\min \{k, nk^{-1}\}$ \\
				\cline{2-3} 
				&	MISDP \eqref{sdp_one_stronger}& $k, nk^{-1} \}$ \\
				\cline{2-3}
				&	MILP \eqref{spca_milp}& $\min\{k(\sqrt{d}/2+1/2),nk^{-1}\sqrt{d} + (n-k) (\sqrt{d}/2+1/2) \}$ \\
				\hline
				\multirow{4}{4.5em}{R1-SSVD}	&MISDP \eqref{sdp_two_eq_svd} & $ \sqrt{mnk_1^{-1}k_2^{-1}}$ \\
				\cline{2-3}
				&MISDP \eqref{svd_sdp_one} & $\min \{\sqrt{k_1k_2}, \sqrt{mnk_1^{-1}k_2^{-1}}\}$ \\
				\cline{2-3}
				& 	\multirow{2}{5.5em}{MILP \eqref{eq:spca_milp_svd}} & $\sqrt{mnk_1^{-1}k_2^{-1}} [\min\{(k_1+k_2)(\sqrt{d}/2+1/2),$\\
				& & $mnk_1^{-1}k_2^{-1}\sqrt{d} + (m+n-k_1-k_2) (\sqrt{d}/2+1/2)\}-1] $\\
				\hline
				\multirow{1}{4.5em}{SFPCA\tnote{2}} 	& MISDP \eqref{sfpca_sdp} & -- \\
				\hline
				\hline
				\multicolumn{1}{|c|}{\textbf{Problem}}	&\multicolumn{1}{c|}{\textbf{Approximation Algorithm}} & \multicolumn{1}{c|}{\textbf{Approximation Ratio\tnote{3}}}\\
				\hline
			\multirow{2}{4.5em}{SPCA} & Greedy \Cref{alg:greedy} & $k^{-1}$\\
				\cline{2-3}
		&Local Search \Cref{alg:localsearch} & $k^{-1}$ \\
			\hline
				\multirow{3}{4.5em}{R1-SSVD}	&Truncation algorithm & $\max\{\sqrt{k_1^{-1}}, \sqrt{k_2^{-1}},\sqrt{k_1k_2m^{-1}n^{-1}} \} $ \\
					\cline{2-3}
					& Greedy \Cref{alg:svd_greedy} & $\sqrt{k_1^{-1}k_2^{-1}}$\\
					\cline{2-3}
					& Local Search \Cref{alg:svd_localsearch} & $\sqrt{k_1^{-1}k_2^{-1}}$\\
				\hline
			\end{tabular}%
			\begin{tablenotes}
				\item[1] Optimality Gap is the ratio between the continuous relaxation value and the optimal one;
				\item[2] The formulation \eqref{sfpca_sdp} provides an upper bound for general SFPCA and becomes exact when there are only two groups;
			 		\item[3] Approximation Ratio denotes the ratio between the objective value of an approximation algorithm and the optimal one.
			\end{tablenotes} 
		\end{threeparttable}
	\end{table}
	
	\noindent \textit{Organization:} The remainder of this paper is organized as follows. Sections \ref{sec:sdpone} and \ref{sec:sdpsecond} develop two MISDP formulations for SPCA and prove the optimality gaps of their continuous relaxation values. Section \ref{sec:MILP} investigates an approximate MILP, which can be arbitrarily close to the optimal value of SPCA, and proves the optimality gap of its continuous relaxation value. Section~\ref{sec:approx_alg} introduces and analyzes two approximation algorithms. Section \ref{sec:computation} conducts a numerical study to demonstrate the efficiency and the solution quality of our proposed formulations and algorithms. Sections \ref{sec_svd} and \ref{sec:sfpca} separately extend the analyses to the rank-one sparse SVD (R1-SSVD) and the sparse fair PCA (SFPCA). Finally, conclusion and future directions are exhibited in Section \ref{sec:conclusion}. 
	
	\noindent \textit{Notation:} The following notation is used throughout the paper. We let $\S^n,\S_+^n, \S_{++}^n$ denote set of all the $n\times n$ symmetric real matrices, set of all the $n\times n$ symmetric positive semi-definite matrices, and set of all the $n\times n$ symmetric positive definite matrices, respectively. We use bold lower-case letters (e.g., $\bm{x}$) and bold upper-case letters (e.g., $\bm{X}$) to denote vectors and matrices, respectively, and use corresponding non-bold letters (e.g., $x_i, X_{ij}$) to denote their components. {We use $\bm{0}$ to denote the zero vector and $\bm{1}$ to denote the all-ones vector.} We use $\lceil \cdot \rceil$ as a ceil function. We let $\Re^n_+$ denote the set of all the $n$ dimensional nonnegative vectors and let $\Re^n_{++}$ denote the set of all the $n$ dimensional positive vectors. Given a positive integer $n$ and an integer $s\le n$, we let $[n]:=\{1,2,\cdots, n\}$ and let $[s,n]:=\{s,s+1,\cdots, n\}$. We let $\bm{I}_n$ denote the $n \times n$ identity matrix and let $\bm{e}_i$ denote its $i$-th column vector. Given a set $S$ and an integer $k$, we let $|S|$ denote its cardinality and $\binom{S}{k}$ denote the collection of all the size-$k$ subsets out of $S$. Given an $m \times n$ matrix $\bm{A}$ and two sets $S\in [m]$, $T\in [n]$, we let $\bm{A}_{S,T}$ denote a submatrix of $\bm{A}$ with rows and columns indexed by sets $S, T$, respectively and let $\bm{A}_S$ denote a submatrix of $\bm{A}$ with columns from the set $S$. 
	Given a vector $\bm{x} \in \Re^n$, we let $\Diag(\bm{x})$ denote the diagonal matrix with diagonal elements $x_1,\cdots, x_n$, and {let $\supp(\bm{x})$ denote the support of $\bm{x}$.} 
	Given a square symmetric matrix $\bm{A}$, let $\diag(\bm{A})$ denote the vector of diagonal entries of $\bm{A}$, and let $\lambda_{\min}(\bm{A}),\lambda_{\max}(\bm{A})$ denote the smallest and largest eigenvalues of $\bm{A}$, respectively. Given a non-square matrix $\bm A$, let $\sigma_{\max}(\bm A)$ denote the largest singular value. Additional notation will be introduced later as needed.

	\section{Exact MISDP Formulation (I)} \label{sec:sdpone}
	In this section, we derive an equivalent mixed-integer semi-definite programming (MISDP) formulation for SPCA based on the spectral decomposition and disjunctive programming techniques. 

	To begin with, for each $i\in [n]$, we let the binary variable $z_i=1$ if the $i$-th feature is selected, and 0, otherwise. Linearizing the zero-norm constraint using binary vector $\bm z$, then SPCA \eqref{spca} can be equivalently formulated as a following nonconvex mixed-integer quadratic program:
	\begin{align}\label{spca2}
	\text{\rm (SPCA)} \quad w^{*} := \max_{\bm{x} \in \Re^n,\bm{z}\in Z} \bigg \{\bm{x}^{\top}\bm{A}\bm{x}: {||\bm{x}||_{2}=1},|x_i|\leq z_i,\forall i \in [n] \bigg \},
	\end{align}
	where we let cardinality set $Z$ denote the feasible region of $\bm z$, i.e., 
	\[Z=\bigg\{ \bm{z}\in \{0,1\}^n: \sum_{i \in [n]}z_i = k\bigg\}.\]
	For SPCA \eqref{spca2}, we note that (i) the binary vector $\bm{z}$ is of vital importance and its associated feasible region $Z$ will be used throughout this paper for two MISDPs and one MILP, and (ii) the derivations of all the three mixed-integer formulations originate from the naive SPCA \eqref{spca2}.
	
	
	\subsection{Spectral Reformulation}
	
	We observe that given a size-$k$ subset of features (i.e., the support of the binary vector $\bm{z}$ in formulation \eqref{spca2} is specified), the SPCA \eqref{spca2} is equivalent to finding the largest eigenvalue of the corresponding principal submatrix of $\bm A$. This fact inspires us to propose three equivalent mixed-integer convex programs for SPCA \eqref{spca2} . This observation is summarized below.
	\begin{restatable}{lemma}{lemmaspca} \label{lemspca}
				For a symmetric matrix $\bm{A}\in \S^n$ and a size-$k$ set $S\subseteq [n]$, the followings must hold:
		\begin{enumerate}[(i)]
			\item $\max_{\bm{x} \in \Re^n} \left \{\bm{x}^{\top}\bm{A}\bm{x}: {||\bm{x}||_{2}=1}, x_i=0, \forall i \notin S \right \} = \lambda_{\max}(\bm{A}_{S,S})$,
			\item $\max_{\bm{X} \in \S_+^k} \left \{ \tr(\bm{A}_{S,S}\bm{X}) : \tr(\bm{X})=1 \right \} = \lambda_{\max}(\bm A_{S,S})$, and
			\item If matrix $\bm A$ is positive semi-definite, then $\lambda_{\max}(\bm{A}_{S,S}) = \lambda_{\max}(\sum_{i\in S}\bm c_i \bm c_i^{\top})$, where $\bm{A} = \bm{C}^{\top}\bm{C}$, $\bm C\in \Re^{d\times n}$ denotes the Cholesky factorization matrix of $\bm{A}$, $d$ is the rank of $\bm{A}$, and $\bm c_i \in \Re^d$ denotes $i$-th column vector of $\bm{C}$ for each $i\in [n]$.
		\end{enumerate}
	\end{restatable}
\begin{proof}
			See Appendix \ref{proof:lemspca}. \qed
\end{proof}
	
	
	
	The results in \Cref{lemspca} are crucial to this paper and allow us to derive the exact mixed-integer convex programs of SPCA. Specifically, we remark that: Part (i) of \Cref{lemspca} reduces SPCA to selecting the best size-$k$ principal submatrix of $\bm A$ to achieve the maximum largest eigenvalue, which establishes a combinatorial formulation of SPCA;
	Part (ii) of \Cref{lemspca} shows that SDP relaxation of the largest eigenvalue problem by dropping the rank-one constraint is exact and inspires us to develop two MISDP formulations for SPCA; and since the covariance matrix used in SPCA is always positive semi-definite, the identity in Part (iii) of \Cref{lemspca} suggests an alternative way of formulating SPCA using Cholesky decomposition, which motivates us to derive an exact MISDP formulation in this section and an MILP in a later section.
	
	
	According to Part (i) in \Cref{lemspca}, introducing a subset $S$, a natural combinatorial reformulation of SPCA \eqref{spca} is defined as:
	\begin{align} \label{spcacom}
	w^{*} := \max_{S} \left \{{\lambda_{\max} (\bm A_{S, S})}: |S|=k, S\subseteq [n] \right \}.
	\end{align}
	%
	By computing the Cholesky factorization of $\bm A= \bm C^{\top} \bm C$ with $\bm C \in \Re^{d \times n}$ and $d$ denoting the rank of $\bm A$, then the identity in Part (iii) in \Cref{lemspca} recasts the objective function of SPCA \eqref{spcacom} as below:
	\begin{align} \label{spcacom2}
	w^{*} := \max_{S} \bigg \{\lambda_{\max}\bigg(\sum_{i\in S}\bm c_i \bm c_i^{\top}\bigg): |S|=k, S\subseteq [n] \bigg \}.
	\end{align}
	Recall that for each $i\in [n]$, binary variable $z_i=1$ if $i$th feature (i.e., column $\bm c_i$) is selected, and 0, otherwise. Therefore, SPCA \eqref{spcacom2} can be further reformulated as 
	\begin{align}\label{spcacom3}
	w^{*} := \max_{\bm z\in Z} \bigg \{{\lambda_{\max} \bigg(\sum_{i\in [n]}z_i \bm c_i \bm c_i^{\top}\bigg)} \bigg \}.
	\end{align}
	
	The above formulation involves with concave objective function but it is a maximization problem, which will cause much trouble. Fortunately, the result in Part (ii) of \Cref{lemspca} and the reformulation technique from disjunctive programming \cite{balas1975disjunctive} motivate us to convert SPCA \eqref{spcacom3} to an equivalent MISDP, which is shown as below.
	\begin{theorem}\label{thm_model2}
		The SPCA \eqref{spca2} admits an equivalent MISDP formulation
		\begin{align} \label{sdp_two_eq}
	\text{\rm (SPCA)} \quad	w^*:=\max_{\begin{subarray}{c}
			\bm{z}\in Z, \\
			\bm X,\bm W_1 , \cdots , \bm W_n \in \S_+^d
			\end{subarray}} \Bigg\{ \sum_{i \in [n]} \bm{c}_i^{\top} \bm{W}_i \bm{c}_i : &\tr(\bm{X}) =1, \bm{X} \succeq \bm{W}_i, \tr(\bm{W}_i) = z_i, \forall i \in [n]\Bigg\}.
		\end{align} 
	\end{theorem}
	\begin{proof}
		According to Part (ii) in \Cref{lemspca}, the largest eigenvalue of a symmetric matrix can be equivalently reformulated as an SDP, thus by introducing a positive semi-definite matrix variable $\bm{X}\in \S_+^d$, SPCA \eqref{spcacom3} can be represented as
		\begin{align}
		w^*:=\max_{\bm{z}\in Z, \bm{X}\in \S_+^d} \Bigg\{ \sum_{i \in [n]} z_i \bm{c}_i^{\top} \bm{X} \bm{c}_i : \tr(\bm{X})=1 \Bigg\}, \label{sdp_two}
		\end{align}
		where the objective function comes from the identity $\tr(\bm{c}_i \bm c_i^{\top}\bm{X})=\bm{c}_i^{\top} \bm{X} \bm{c}_i$ for each $i\in [n]$.
		
		
		
		In SPCA \eqref{sdp_two}, the objective function contains bilinear terms $\{z_i\bm{X}\}_{i\in [n]}$.
		To further convexify them, we create two copies of the matrix variable $\bm{X}$, denoting by $\bm{W}_{i1},\bm{W}_{i2}$ for each $i\in [n]$ and one of them will be equal to $\bm{X}$ depending on the value of binary variable $z_i$. Specifically, SPCA \eqref{sdp_two} now becomes
		\begin{align*} 
		w^*:=\max_{\bm{z}\in Z, \bm{X}, \bm{W}_{i1}, \bm{W}_{i2}\in \S_+^d} \Bigg\{ &\sum_{i \in [n]} \bm{c}_i^{\top} \bm{W}_{i1} \bm{c}_i : \bm{X} =\bm{W}_{i1} + \bm{W}_{i2}, \forall i \in[n], \tr(\bm{X})=1, \notag\\
		&\tr(\bm{W}_{i1}) = z_i,\tr(\bm{W}_{i2}) = 1-z_i, \forall i \in [n]\Bigg\}.
		\end{align*}
		Above, the matrix variables $\{\bm{W}_{i2}\}_{i\in [n]}$ are redundant and can be replaced by inequality $\bm{X} \succeq \bm{W}_i$ for each $i\in [n]$. Thus, we arrive at the equivalent reformulation \eqref{spcacom2} for SPCA. 
		\qed
	\end{proof}
	
\Cref{thm_model2} presents the first equivalent MISDP formulation \eqref{sdp_two_eq} to SPCA. 
	The resulting formulation \eqref{sdp_two_eq} has several interesting properties: (i) it can be directly solved via exact MISDP solvers such as YALMIP; (ii) matrix variables $\bm{X}$ and $\{\bm{W}_i\}_{i\in [n]}$ have dimension of $d\times d$, where $d$ is the rank of matrix $\bm{A}$. Thus, the size of SPCA \eqref{sdp_two_eq} can be further reduced if the covariance matrix $\bm A$ is low-rank; and (iii) the binary variables $\bm{z}$ can be separated from the other variables, so one can apply the Benders decomposition to solving the SPCA \eqref{sdp_two_eq}. This result will be elaborated with more details in the next subsection.
	
	For large-scale instances, computing the continuous relaxation values of the SPCA \eqref{sdp_two_eq} provides us an upper bound to the optimal value or can be useful to check the quality of different heuristics.
	In the following, we show that the continuous relaxation value of SPCA \eqref{sdp_two_eq} is not too far away from the optimal value $w^*$.
	First, let $\overline{w}_1$ denote the continuous relaxation value, i.e.,
	\begin{align} \label{sdp_two_eq_re}
	\overline{w}_1:=\max_{\begin{subarray}{c}
		\bm{z}\in \overline{Z}, \\
		\bm X,\bm W_1 , \cdots , \bm W_n \in \S_+^d
		\end{subarray}} \Bigg\{ \sum_{i \in [n]} \bm{c}_i^{\top} \bm{W}_i \bm{c}_i : &\tr(\bm{X}) =1, \bm{X} \succeq \bm{W}_i, \tr(\bm{W}_i) = z_i, \forall i \in [n]\Bigg\},
	\end{align} 
	where we let $\overline{Z}$ denote the continuous relaxation of set $Z$, i.e.,
	\[\overline{Z}=\bigg\{ \bm{z}\in [0,1]^n: \sum_{i \in [n]}z_i = k\bigg\}.\]
	
	\begin{theorem}\label{thm_model2_relax} 
		The continuous relaxation value $\overline{w}_1$ of formulation \eqref{sdp_two_eq} achieves a $\min\{k, n/k\}$ optimality gap of SPCA, i.e.,
		$$w^* \le \overline{w}_1 \le \min\{k, n/k\} w^*.$$
	\end{theorem}
	\begin{proof}
		It is obvious that $w^* \le \overline{w}_1$ since the feasible region of continuous relaxation \eqref{sdp_two_eq_re} includes the original decision space. Thus, it remains to show that (i) $\overline{w}_1 \le k w^*$ and (ii) $\overline{w}_1 \le n/k w^*$.
		
		\begin{itemize}
			\item[Part (i) $\overline{w}_1 \le k w^*$.] 
			For any feasible solution $(\bm{z},\bm{X},\{\bm{W}_i\}_{i\in [n]})$ to problem \eqref{sdp_two_eq_re}, we must have 
			\begin{align*}
			\sum_{i \in [n]} \bm c_i^{\top} \bm W_i \bm c_i \leq \sum_{i \in [n]} \bm c_i^{\top} \bm c_i \tr(\bm W_i) = \sum_{i\in [n]} z_i \bm c_i^{\top}\bm c_i \leq \sum_{i\in [n]} z_i w^* =k w^*,
			\end{align*}
			where the first inequality is due to the fact that the trace of the product of two symmetric positive semi-definite matrices is no larger than the product of the traces of these two matrices \citep{coope1994matrix}, the first equality is from $\tr(\bm W_i)=z_i$ for each $i\in [n]$, the second inequality is because 
			\[\bm c_i^{\top}\bm c_i=\lambda_{\max}\left(\bm c_i \bm c_i^{\top}\right)\leq \max_{S\subseteq [n]:|S|=k}\lambda_{\max}\bigg(\sum_{j\in S}\bm c_j \bm c_j^{\top}\bigg):=w^*,\]
			and the second equality is due to $\sum_{i\in [n]} z_i=k$.
			
			\item[Part (ii) $\overline{w}_1 \le n/k w^*$.]Similarly, given any feasible solution $(\bm{z},\bm{X},\{\bm{W}_i\}_{i\in [n]})$ of continuous relaxation \eqref{sdp_two_eq_re}, we must have
			\begin{align*}
			\sum_{i \in [n]} \bm c_i^{\top} \bm W_i \bm c_i \le \sum_{i\in [n]} \bm c_i^{\top} \bm X \bm c_i=\frac{1}{{{n-1}\choose{k-1}}}\sum_{S\in {{[n]}\choose{k}}} \sum_{i\in S}\bm c_i^{\top} \bm X \bm c_i \le \frac{{{n}\choose{k}}}{{{n-1}\choose{k-1}}}w^*= \frac{n}{k} w^*,
			\end{align*}
			where the first inequality is because $\bm W_i\succeq \bm{X}$ and the second one is from Part (ii) in \Cref{lemspca}.\qed
		\end{itemize}
	\end{proof}
	
	\Cref{thm_model2_relax} shows that the continuous relaxation value of formulation \eqref{sdp_two_eq_re} is at most $\min\{k, n/k\}$ away from the true optimal value of SPCA \eqref{sdp_two_eq}, implying that if $k\rightarrow 1$ or $k\rightarrow n$, then the continuous relaxation value $\overline{w}_1$ is very close to the true optimal value $w^*$, which is consistent with the numerical study in \Cref{sec:computation}.

	\subsection{Solving SPCA \eqref{sdp_two_eq} and SDP Relaxation \eqref{sdp_two_eq_re}: Benders Decomposition} \label{branch}
	It has been recognized that large-scale SDPs are challenging to solve, so is the MISDP \eqref{sdp_two_eq}. In this subsection, we apply the Benders decomposition \citep{benders62,geoffrion1972generalized} to the proposed MISDP \eqref{sdp_two_eq}, 
which can be further integrated into the branch and cut framework.
By relaxing the binary vector $\bm{z}$ to be continuous, the Benders Decomposition recasts the continuous SDP relaxation \eqref{sdp_two_eq_re} as a maximin saddle point problem, which enables the adoption of the efficient subgradient method.
	
%

The main idea of Benders decomposition is to decompose SPCA \eqref{sdp_two_eq} into two stages: first, the master problem is a pure integer maximization problem over $\bm{z}$, and second, given a feasible $\bm z \in Z$, the subproblem is to maximize over the remaining variables $(\bm X, \{\bm W_i\}_{i\in[n]})$.
Thus, by separating the binary variables, we rewrite the SPCA \eqref{sdp_two_eq} as
	\begin{align} \label{sdp_two_eq_h1}
	w^*:=\max_{\bm{z}\in Z} H_1(\bm{z}):=\max_{
		\bm X,\bm W_1 , \cdots , \bm W_d \in \S_+^d} \Bigg\{ \sum_{i \in [n]} \bm{c}_i^{\top} \bm{W}_i \bm{c}_i : &\tr(\bm{X}) =1, \bm{X} \succeq \bm{W}_i, \tr(\bm{W}_i) = z_i, \forall i \in [n]\Bigg\}.
	\end{align}
	
Benders decomposition is of particular interest when the subproblem $H_1(\bm{z})$ for any $\bm z \in Z$ is easy to compute, which is, unfortunately, not the case. Therefore,	it is desirable if we can specify the function $H_1(\bm{z})$ for any given $\bm{z} \in Z$ in an efficient way. Surprisingly, invoking Part(ii) in \Cref{lemspca}, the strong duality of inner SDP maximization problem in \eqref{sdp_two_eq_h1} holds and the obtained dual problem admits a closed-form solution for any binary variables $\bm z \in Z$, which enables the subproblem to generate valid inequalities to the master problem efficiently. The results are shown below.
	
	\begin{restatable}{proposition}{thmHone}\label{thm_H1}
For the	function $H_1(\bm{z})$ defined in \eqref{sdp_two_eq_h1}, we have 
		\begin{enumerate}[(i)]
	\item 	For any $\bm z\in \overline{Z}$, function $H_1(\bm{z})$ is equivalent to
	\begin{align}\label{eq_discrbi_H14}
	H_1(\bm{z})=\min_{ \bm{\mu},\bm Q_1,\cdots,\bm Q_n\in \S_+^d} &\bigg\{\lambda_{\max}\bigg(\sum_{i\in [n]} \bm Q_i \bigg) + \sum_{i \in [n]} \mu_i z_i: \bm{c}_i\bm{c}_i^{\top} \preceq \bm{Q}_i + \mu_{i} \bm{I}_d , 0\le \mu_{i}\le \|\bm{c}_i\|_2^2 , \forall i \in [n] \bigg\},
	\end{align}
	which is concave in $\bm{z}$.
\item For any binary $\bm z\in {Z}$, an optimal solution to problem \eqref{eq_discrbi_H14} is $\mu^*_i=0$ if $z_i = 1$ and $\|\bm{c}_i\|_2^2$, otherwise, and $\bm{Q}^*_i:=(1-\mu_i^*/\|\bm{c}_i\|_2^2) \bm{c}_i\bm{c}_i^\top$ for each $i\in [n]$.
		\end{enumerate}
	\end{restatable}
\begin{proof}
	See Appendix \ref{proof_thm_H1}. \qed
\end{proof}

The Part (ii) of \Cref{thm_H1} shows that given a solution $\bm z \in Z$ with its support $S$, the optimal value to \eqref{eq_discrbi_H14} is equal to
	\[H_1(\bm z) = \lambda_{\max}\bigg(\sum_{i\in S}\bm c_i \bm c_i^{\top}\bigg) + \sum_{i\in [n]\setminus S} ||\bm c_i||^2_2,\]
which leads to an equivalent reformulation of SPCA \eqref{sdp_two_eq_h1} as
\begin{equation}\label{spca_exp}
w^*=\max_{\bm{z}\in Z} \bigg\{w: w\leq \lambda_{\max}(\bm{A}_{SS})+\sum_{i\in [n]\setminus S}\|\bm{c}_i\|_2^2z_i, \forall S\subseteq [n]: |S|=k\bigg\}.
\end{equation}
Above, for any mixed binary solution $(\hat{\bm{z}},\hat{w})\in Z\times \Re$, the most violated constraint is
\[w\leq \lambda_{\max}(\bm{A}_{\hat S\hat S})+\sum_{i\in [n]\setminus \hat S}\|\bm{c}_i\|_2^2z_i,\]
where set $\hat S:=\{i\in [n]: \hat z_i=1\}$ denotes the support of $\hat{\bm{z}}$.
We remark that the exact branch and cut approach to solve SPCA \eqref{spca_exp} using \textit{callback} functions will benefit from these closed-form valid inequalities.
	
%

Note that by relaxing the binary variables to be continuous, the relaxed problem \eqref{sdp_two_eq_h1} is equivalent to the SDP relaxation \eqref{sdp_two_eq_re}.
However, given $z \in \overline{Z}$, the dual representation of function $H_1(\bm{z})$ in \eqref{eq_discrbi_H14} is still a difficult SDP. Motivated by Part (ii) in \Cref{thm_H1}, we propose a more efficient upper bound $\overline{H}_1(\bm{z})$ than ${H}_1(\bm{z})$ by letting $\bm{Q}_i:=(1-\mu_i/\|\bm{c}_i\|_2^2) \bm{c}_i\bm{c}_i^\top$ for each $i\in [n]$ to problem \eqref{eq_discrbi_H14}. In the next theorem, we show that the relaxed $\overline{H}_1(\bm{z})$ becomes exact for any binary vector $\bm z\in Z$ and the resulting upper bound of SPCA also achieves a $\min\{k, n/k\}$ optimality gap.	

	\begin{theorem}\label{cor_hat_H_1}The following results hold for the relaxed function $\overline{H}_1(\bm{z})$:
		\begin{enumerate}[(i)]
			\item For any $\bm z\in \overline{Z}$, function $H_1(\bm{z})$ is upper bounded by
			\begin{align}\label{eq_discrbi_H14_hat}
			\overline{H}_1(\bm{z})=\min_{ \bm{\mu}} \bigg\{\lambda_{\max}\bigg(\sum_{i\in [n]} (1-\mu_i/\|\bm{c}_i\|_2^2)\bm c_i \bm c_i^{\top} \bigg) + \sum_{i \in [n]} \mu_i z_i: 0\le \mu_{i}\le \|\bm{c}_i\|_2^2 , \forall i \in [n] \bigg\};\end{align}
			\item If $\bm{z}\in Z$, then $H_1(\bm{z})=\overline{H}_1(\bm{z})=\lambda_{\max}(\sum_{i\in [n]}z_i \bm{c}_i\bm{c}_i^\top)$; and
			\item The continuous relaxation value of SPCA
			\begin{align}\label{eq_overline_w_2}
			\overline{w}_2=\max_{\bm{z}\in \overline{Z}}\overline{H}_1(\bm{z})
			\end{align}
	achieves a $\min\{k, n/k\}$ optimality gap of SPCA, i.e.,
			$w^* \le \overline{w}_1\leq \overline{w}_2 \le \min\{k, n/k\} w^*,$
			where $\overline{w}_1$ is defined in \eqref{sdp_two_eq_re}.
		\end{enumerate}
	\end{theorem}
	\begin{proof}
		\begin{enumerate}[(i)]
			\item The conclusion follows by choosing a feasible $\bm{Q}_i:=(1-\mu_i/\|\bm{c}_i\|_2^2) \bm{c}_i\bm{c}_i^\top$ for each $i\in [n]$ in the representation \eqref{eq_discrbi_H14}.
			
			\item For any $\bm{z}\in Z$, we derive from Part (ii) in \Cref{thm_H1} that $\overline{H}_1(\bm{z})\geq \lambda_{\max}(\sum_{i\in [n]}z_i \bm{c}_i\bm{c}_i^\top)$. Thus, it is sufficient to show that $\overline{H}_1(\bm{z})\leq \lambda_{\max}(\sum_{i\in [n]}z_i \bm{c}_i\bm{c}_i^\top)$. Indeed, this can be done simply by letting $\mu_i=0$ if $z_i=0$, and $||\bm c_i||_2^2$, otherwise in \eqref{eq_discrbi_H14_hat}.	
			
			\item By the proof of \Cref{thm_model2_relax}, to obtain the same optimality gap for \eqref{eq_overline_w_2} as SDP \eqref{sdp_two_eq_re}, we need to show that $\overline{H}_1(\bm{z})\leq \sum_{i\in [n]}z_i\bm{c}_i^\top\bm{c}_i$ and $\overline{H}_1(\bm{z})\leq \lambda_{\max}(\bm A)=\lambda_{\max}(\sum_{i\in [n]}\bm{c}_i\bm{c}_i^\top)$ for any $\bm{z}\in \overline{Z}$.
			
			We must have $\overline{H}_1(\bm{z})\leq \sum_{i\in [n]}z_i\bm{c}_i^\top\bm{c}_i$ by by letting $
			\mu_i=\bm{c}_i^\top\bm{c}_i$ for all $i\in [n]$ in \eqref{eq_discrbi_H14_hat}.
			
			We also have $\overline{H}_1(\bm{z})\leq \lambda_{\max}(\bm A)=\lambda_{\max}(\sum_{i\in [n]}\bm{c}_i\bm{c}_i^\top)$ by letting $\mu_i=0$ for all $i\in [n]$ in \eqref{eq_discrbi_H14_hat}.
			
			Then the rest of the proof follows directly from that of \Cref{thm_model2_relax} and is thus omitted.
		\end{enumerate}\qed
	\end{proof}
	We remark that: (i) Compared to $H_1(\bm{z})$, 	function $\overline{H}_1(\bm{z})$ in \eqref{eq_discrbi_H14_hat} only involves an $n$-dimensional variable $\bm \mu$. 
The resulting relaxation \eqref{eq_overline_w_2} of SPCA can be viewed as a conventional saddle problem so we apply the subgradient method with convergence rate of $O(1/{T})$ to the search for optimal solutions (see, e.g., \cite{nedic2009subgradient}), which offers an efficient way to generate an upper bound of SPCA in \Cref{sec:computation};
	(ii) On the other hand, the continuous relaxation value $\overline{w}_1=\max_{\bm{z}\in \overline{Z}}H_1(\bm{z})$ tends to be stronger than $\overline{w}_2$ in \eqref{eq_overline_w_2}. Thus, it is a tradeoff between computational effort and a better upper bound; (iii) Surprisingly, both bounds $\overline{w}_1,\overline{w}_2$ achieve the same optimality gap of SPCA. This implies that there might be room to improve the analysis of optimality gap in \Cref{thm_model2_relax}. We leave this to interested readers; and (iv) more importantly, when $\bm{z}\in Z$ is binary, both problems \eqref{eq_discrbi_H14} and \eqref{eq_discrbi_H14_hat} have closed-form results, which are very helpful for using the branch and cut method. 

	\section{Exact MISDP Formulation (II)}\label{sec:sdpsecond}
	The MISDP formulation \eqref{sdp_two_eq} developed for SPCA in the previous section mainly are inspired from Part(ii) and Part(iii) in \Cref{lemspca}. In this section, we will propose another exact MISDP reformulation of SPCA using Part(i) and Part(ii) in \Cref{lemspca}. Similarly, we will present the optimality gap of the corresponding SDP relaxation to demonstrate the strength of the second formulation. 
It is worthy of noting that the proposed MISDP \eqref{sdp_two_eq} requires the positive semi-definiteness of matrix $\bm{A}$ as it is built on Cholesky decomposition of $\bm{A}$, but the result in this section is more general and holds even matrix $\bm{A}$ is not positive semi-definite.
	
	\subsection{A Naive Exact MISDP Formulation}
	
	We first establish a naive exact MISDP formulation of SPCA \eqref{spca2} based on Part (ii) in \Cref{lemspca}, and the resulting continuous relaxation value is equal to $\lambda_{\max}(\bm{A})$.
	
	\begin{restatable}{proposition}{sdptwoprop}\label{them_eq_pca}
	 The SPCA \eqref{spca2} admits the following MISDP formulation:
		\begin{align} \label{sdp_one}
\text{\rm (SPCA)}	\quad	w^{*} := \max_{\bm{z} \in Z, \bm{X}\in \S^n_+} \bigg \{ \tr(\bm{A}\bm{X}) : \tr(\bm{X})=1, X_{ii} \le z_{i}, \forall i \in [n] \bigg \}.
		\end{align}
		and its continuous relaxation value is equal to $\lambda_{\max}(\bm{A})$.
	\end{restatable}
	\begin{proof}
		See Appendix \ref{proof_prop2}. \qed
	\end{proof}
	The SPCA formulation \eqref{sdp_one} can be also found in \cite{gally2016computing}. However, our proof is quite different and shorter, since it does not involve sophisticated extreme point characterization of SDPs. 
	Although the MISDP \eqref{sdp_one} is equivalent to SPCA \eqref{spca2}, the fact that its continuous relaxation value is equal to $\lambda_{\max}(\bm{A})$ demonstrates that it might be a weak formulation. This motivates us to further strengthen the formulation \eqref{sdp_one} by adding valid inequalities in the next subsection. 
	
	\subsection{A Stronger Reformulation with Two Valid Inequalities}
	In this subsection, we first propose two valid inequalities for SPCA \eqref{sdp_one} and derive the optimality gap of its continuous relaxation value of the improved formulation. 
	
	After examining different types of valid inequalities, we propose the following two types of valid inequalities for the SPCA formulation \eqref{sdp_one}.
	\begin{restatable}{lemma}{lemineq}\label{lem_ineq}
		The following two inequalities are valid to SPCA \eqref{sdp_one}
		\begin{enumerate}[(i)]
			\item $\sum_{j\in [n]} X_{ij}^2 \le X_{ii} z_i $ for all $i \in [n]$; and
			\item $\left(\sum_{j \in [n]} | X_{ij}| \right)^2 \le k X_{ii}z_i$ for all $i \in [n]$.
		\end{enumerate}
	\end{restatable}
	\begin{proof}
		See Appendix \ref{proof_lem_ineq}. \qed
	\end{proof}
	We make the following remarks about \Cref{lem_ineq}.
	\begin{enumerate}[(i)]
		\item Many other valid inequalities are dominated by the two types of valid inequalities in \Cref{lem_ineq} such as
		\begin{align*}
		|X_{ij}|\leq z_i, X_{ij}^2\leq X_{ii}z_j,
		X_{ij}^2\leq z_iz_j, \forall i,j\in [n];
		\end{align*}
		\item Note that the two types of valid inequalities are both second order conic (see e.g., \cite{ben2001lectures}), and thus can be embedded into SDP solvers such as MOSEK, SDPT3; and
		
		\item We further observe that the inequality $ X_{ii}\leq z_i$ in \eqref{sdp_one} is dominated by the first type of inequalities with the facts that $ X_{ii}^{2}+\sum_{j\in [n]\setminus\{i\}} X_{ij}^2 \le X_{ii} z_i $ and $X_{ii}\geq 0$ for each $i\in [n]$.
	\end{enumerate}
	
	The results in \Cref{lem_ineq} together with \Cref{them_eq_pca} give rise to a stronger MISDP of SPCA than formulation \eqref{sdp_one}, which is summarized below.
	\begin{theorem} \label{them_eq_pca_stronger}
		The SPCA \eqref{spca2} can reduce to following stronger MISDP formulation:
		\begin{align} \label{sdp_one_stronger}
\text{\rm (SPCA)}	\	w^{*} := \max_{\bm{z} \in Z, \bm{X}\in \S^n_+} &\bigg \{ \tr(\bm{A}\bm{X}) : \tr(\bm{X})=1,\sum_{j\in [n]} X_{ij}^2 \le X_{ii} z_i, \bigg(\sum_{j \in [n]} | X_{ij}| \bigg)^2 \le k X_{ii}z_i,\forall i \in [n] \bigg \}.
		\end{align}
	\end{theorem}
	Let $\overline{w}_3$ denote the continuous relaxation value of SPCA formulation \eqref{sdp_one_stronger}, i.e.,
	\begin{align} \label{sdp_one_stronger_rel}
	\overline{w}_3 := \max_{\bm{z} \in \overline{Z}, \bm{X}\in \S^n_+} &\bigg \{ \tr(\bm{A}\bm{X}) : \tr(\bm{X})=1, \sum_{j\in [n]} X_{ij}^2 \le X_{ii} z_i, \bigg(\sum_{j \in [n]} | X_{ij}| \bigg)^2 \le k X_{ii}z_i,\forall i \in [n] \bigg \}.
	\end{align}
	
	Clearly, we have $\lambda_{\max}(\bm A)\geq \overline{w}_3$. We are going to prove that the continuous relaxation value can be even stronger than a well-known SDP upper bound for SPCA \eqref{spca2} introduced by \citet{d2005direct}, denoted by $\overline{w}_4$, that has been widely used for solving SPCA in literature. The upper bound from \cite{d2005direct} comes to the following formulation
	\begin{align} \label{eq_sdp_d2005direct}
	\overline{w}_4 := \max_{ \bm{X}\in \S^n_+} &\bigg \{ \tr(\bm{A}\bm{X}) : \tr(\bm{X})=1, \sum_{i \in [n]} \sum_{j \in [n]} | X_{ij}| \le k \bigg \}.
	\end{align}
The formal comparison result is shown below.

	\begin{proposition}\label{prop_bound_comparison_model1}
	The upper bounds $\overline{w}_3,\overline{w}_4$ of SPCA defined in \eqref{sdp_one_stronger_rel} and\eqref{eq_sdp_d2005direct}, respectively, satisfy
		$\overline{w}_4\geq \overline{w}_3$, i.e., the continuous relaxations value of the stronger MISDP \eqref{sdp_one_stronger} is stronger than the optimal value of the SDP formulation \eqref{eq_sdp_d2005direct} from \cite{d2005direct}.
	\end{proposition}
	\begin{proof} 
		To show that $\overline{w}_4\geq \overline{w}_3$, it is sufficient to prove that any feasible solution $(\bm{z},\bm{X})$ of the continuous relaxation problem \eqref{sdp_one_stronger_rel}, will satisfy the constraints in the SDP formulation \eqref{eq_sdp_d2005direct}. 
		
		Clearly, we have $\bm{X}\in \S_+^n$ and $\tr(\bm{X})=1$. It remains that $\sum_{i \in [n]} \sum_{j \in [n]} | X_{ij}| \le k$. Indeed, we have
		\begin{align*}
		\sum_{i \in [n]} \sum_{j \in [n]} | X_{ij}| \le \sum_{i \in [n]} \sqrt{k} \sqrt{ X_{ii} z_i} \le \sqrt{k} \sqrt{\sum_{i \in [n]} X_{ii}} \sqrt{\sum_{i \in [n]} z_{i}} = k,
		\end{align*}
		where the first inequality results from type (ii) inequalities in \Cref{lem_ineq}, the second one is due to Cauchy–Schwartz inequality, and the equality is due to $\tr(\bm{X})=1$ and $\sum_{i\in [n]}z_i=k$. \qed
	\end{proof}

	Next, we show that the continuous relaxations value of the stronger MISDP \eqref{sdp_one_stronger} is also quite close to the true value. This phenomenon is more striking in the numerical study.
	\begin{theorem} \label{them_contsdp_one_gap}
		The continuous relaxations value of the stronger MISDP formulation \eqref{sdp_one_stronger} yields a $\min\{k, n/k\}$ optimality gap for SPCA, i..e, 
		$$w^*\le \overline{w}_3 \le \min\{k, n/k\}w^*.$$
	\end{theorem}
	\begin{proof}
		The proof is separated into two parts: (i) $\overline{w}_3 \le k w^*$ and (ii) $\overline{w}_3 \le n/k w^*$.
		\begin{enumerate}[(i)]
			\item $\overline{w}_3 \le k w^*$.
			For any feasible solution $\bm X$ to problem \eqref{sdp_one_stronger_rel}, we have
			\begin{align*}
			\tr(\bm{A} \bm{X}) = \sum_{i \in [n]} \sum_{j \in [n]} A_{ij} X_{ij} \leq \sum_{i \in [n]} \sum_{j \in [n]} | A_{ij}|| X_{ij}|\le w^* \sum_{i \in [n]} \sum_{j \in [n]} | X_{ij}| \le k w^*,
			\end{align*}
			where the first inequality is due to taking the absolute values, the second one is based on the fact that $\max_{i\in [n]}\{A_{i,i}\} \le w^*$ and $\bm |A_{i,j}|\leq \sqrt{A_{i,i}A_{j,j}}\leq w^*$ for each pair $i,j\in [n]$, and the third one can be obtained from the proof of \Cref{prop_bound_comparison_model1}.
			\item $\overline{w}_3 \le n/k w^*$. The proof is similar to the one of \Cref{thm_model2_relax} since $\overline{w}_3\leq \lambda_{\max}(\bm{A})\le n/k w^*$. 		\qed
			
		\end{enumerate}
	\end{proof}
	In general, our two proposed MISDP formulations \eqref{sdp_two_eq} and \eqref{sdp_one_stronger} are not comparable although their continuous relaxations have the same theoretical approximation gap, which will be also illustrated in the numerical study section. The continuous relaxation of the MISDP formulation \eqref{sdp_one_stronger} might be difficult to solve due to lager size of its matrix variables and higher complexity of its constraints.
	In the next subsection, we will discuss Benders decomposition for SPCA \eqref{sdp_one_stronger}, where the subproblem reduces to a second order conic program rather than an SDP.


\subsection{Benders Decomposition}
The decomposition method developed for SPCA \eqref{sdp_one_stronger} in this subsection follows from \Cref{branch}. Therefore, many details will be omitted for brevity. 
Similarly, we decompose the proposed MISDP formulation \eqref{sdp_one_stronger} by a master problem over binary variables $\bm z \in Z$ and a subproblem over the matrix variable $\bm X\in \S_+^n$.
Also, we reformulate SPCA \eqref{sdp_one_stronger} as the following equivalent two-stage optimization problem
\begin{align} \label{sdp_two_eq_h2}
 w^*=\max_{\bm{z}\in Z} H_2(\bm{z}):=\max_{\bm{X}\in \S^n_+} &\bigg \{ \tr(\bm{A}\bm{X}) : \tr(\bm{X})=1,\sum_{j\in [n]} X_{ij}^2 \le X_{ii} z_i, \bigg(\sum_{j \in [n]} | X_{ij}| \bigg)^2 \le k X_{ii}z_i,\forall i \in [n] \bigg \}.
\end{align}
It is favorable to derive an efficient dual formulation of $H_2(\bm{z})$ for any given $\bm{z} \in \overline{Z}$ such that its subgradient can be easily computed. Indeed, invoking Part(ii) in \Cref{lemspca} and dualizing the second order conic constraints, the strong duality of inner maximization over $\bm X$ in \eqref{sdp_two_eq_h2} still holds. The proof is similar to \Cref{thm_H1} and is thus omitted. 
\begin{proposition}\label{thm_H2}
For any $\bm z\in \overline{Z}$, function $H_2(\bm{z})$ is equivalent to
\begin{equation}\label{eq_discrbi_H2}
\begin{aligned}
H_2(\bm{z})=\min_{\bm{\mu}, \bm{\nu}_1,\bm{\nu}_2,\bm{\Lambda},\bm{W}_1,\bm{W}_2,\bm{\beta}} & \lambda_{\max}\left(\bm{A} +\bm{\Lambda} + 1/2 \Diag(\bm{\mu}_1+\bm{\mu}_2+\bm{\nu}_1+\bm{\nu}_2) -\bm{W}_1+\bm{W}_2\right) \\
&+ 1/2 (-\bm{\mu}_1+\bm{\mu}_2)^{\top} \bm{z} + k/2 (-\bm{\nu}_1+\bm{\nu}_2)^{\top} \bm{z}, \\
\text{\rm s.t.}\quad 
&\beta_{i}+(\bm W_{1})_{ij}+(\bm W_{2})_{ij} \le 0 , \forall i \in [n], j \in [n],\\
& \sum_{j \in [n]}\bm \Lambda_{ij}^2 + (\mu_{i1})^2 \le (\mu_{i2})^2, \forall i \in [n], \\
& \beta_{i}^2 + (\nu_{i1})^2 \le (\nu_{i2})^2 , \forall i \in [n],\\
& \bm (W_{1})_{ij} \ge 0, \bm (W_{2})_{ij}\ge 0, \forall i \in [n], \forall j \in [n],\\
& \bm{\nu}_1 , \bm{\nu}_2 \in \Re^n_+, \bm{\Lambda},\bm{W}_1,\bm{W}_2\in \S^n,
\end{aligned}
\end{equation}
which is concave in $\bm{z}$.
\end{proposition}

For the equivalent function ${H}_2(\bm{z})$ derived in \Cref{thm_H2}, we remark that: (i) Note that for any given $\bm{z}\in \overline{Z}$, function ${H}_2(\bm{z})$ can be solved as an second order conic program and escape from the SDP curse. More effectively, it can be solved via many first-order methods (e.g., the subgradient method) since the subgradient is easy to obtain and the projection only involves second order conic constraints; (ii) On the other hand, when we solve the continuous relaxation 
\begin{align}\label{eq_overline_w_4}
\overline{w}_3=\max_{\bm{z}\in \overline{Z}}H_2(\bm{z}),
\end{align}
the subgradient method is also applicable to solve the entire maximin saddle problem with $O(1/{T})$ rate of convergence (see, e.g., \cite{nedic2009subgradient}); (iii) We can warm start the exact branch and cut algorithm by solving the continuous relaxation \eqref{eq_overline_w_4}, and add all the subgradient inequalities into the root relaxed problem.

\section{A Mixed-Integer Linear Program (MILP) for SPCA with Arbitrary Accuracy} \label{sec:MILP} 
The formulations developed in the previous section for solving SPCA either rely on MISDP solvers or customized branch and cut algorithms, which does not leverage existing computational powers of solvers such as CPLEX, Gurobi. In this section, motivated by the SPCA formulation \eqref{spcacom3} and the identity of eigenvalues, we further derive an approximate mixed-integer linear program (MILP) for SPCA with arbitrary accuracy $\epsilon>0$ and $O(n+d+\log(\epsilon^{-1}))$ binary variables. We also prove the optimality gap of its corresponding LP relaxation. The results in this section assume that $\bm{A}$ is positive semi-definite.

\subsection{An MILP Formulation for SPCA}

The difficulty of SPCA \eqref{spcacom3} lies in how to convexify the objective function, i.e., the largest eigenvalue of a symmetric matrix $\bm{A}$. 
In particular, our proposed MISDP formulations stem from the fact that the largest eigenvalue can be formulated as an equivalent SDP problem. Through a different lens, we represent the largest eigenvalue function based on the natural definition of eigenvalues of a matrix, i.e., 
\begin{align*}
\lambda_{\max}(\bm A) =\max_{w,\bm{x}\in \Re^n}\bigg\{w: \bm{A}\bm{x}=w\bm{x},\bm{x}\neq 0\bigg\},
\end{align*}
where $\bm{x}$ denotes an eigenvector and the nonzero constraint rules out the trivial solution $\bm{x}=0$.

This motivates us to recast SPCA formulation \eqref{spcacom3} as the following nonconvex problem
\begin{align}\label{spcacom3_eigen}
w^*=\max_{w,\bm{x}\in \Re^d,\bm{z}\in Z}\bigg\{w: \sum_{i \in [n]}z_i\bm{c}_i\bm{c}_i^\top \bm{x}=w\bm{x},\|\bm{x}\|_{\infty}=1\bigg\},
\end{align}
where $\|\bm{x}\|_{\infty}=1$ also excludes the trivial solution $\bm{x}=\bm{0}$. 

For any given $\bm{z}\in Z$, the nonconvexity of SPCA formulation \eqref{spcacom3_eigen} lies in three aspects: (i) Bilinear terms $\{z_i\bm{x}\}_{i\in [n]}$. They can be easily linearized using the disjunctive programming techniques since vector $\bm{z}$ is binary; (ii) Constraint $\|\bm{x}\|_{\infty}=1$. The nonconvex constraint $\|\bm{x}\|_{\infty}=1$ can be equivalently written as a disjunction with $2d$ sets below
\[\cup_{j\in [d]}\big\{\bm{x} \in \Re^d: x_j=1,\|\bm{x}\|_{\infty}\leq 1 \big\}\cup_{j\in [d]}\big\{\bm{x} \in \Re^d: x_j=-1,\|\bm{x}\|_{\infty}\leq 1 \big\}.\]
Due to the equivalence of $\bm x$ and $-\bm x$ in SPCA \eqref{spcacom3_eigen}, it suffices to only keep first $d$ sets, i.e., $\cup_{j\in [d]}\big\{\bm{x}\in \Re^d: x_j=1,\|\bm{x}\|_{\infty}\leq 1 \big\}$. This disjunction can be equivalently described as an MILP using the results in \cite{balas1975disjunctive}; and (iii) Bilinear term $w\bm{x}$. We can first approximate variable $w$ using binary expansion and then linearize the obtained bilinear terms by the same disjunctive technique as part (i). The resulting MILP formulation is summarized in the following theorem.
\begin{restatable}{theorem}{themmilp}\label{them_milp}
	Given a threshold $\epsilon>0$, the following MILP is $O(\epsilon)$-approximate to SPCA \eqref{spca2}, i.e., $\epsilon\leq \hat w(\epsilon)- w^* \le \epsilon\sqrt{d}$
	\begin{equation}\label{spca_milp}
	\begin{aligned} 
	\hat{w}(\epsilon) := &\max_{\begin{subarray}{c}
		w, \bm{z} \in Z, \bm y, \bm{\alpha},\bm{x}, , \bm \delta,\bm \mu,\bm\sigma
		\end{subarray}} w\\
	\text{\rm s.t.}\quad
	& {\bm{x} = \bm \delta_{i1}+\bm \delta_{i2}, ||\bm \delta_{i1}||_{\infty}\le z_i, ||\bm \delta_{i2}||_{\infty}\le 1- z_i, \forall i \in [n]}, \\
	&{ \bm x = \sum_{j \in [d]} \bm \sigma_{j}, ||\bm \sigma_{j}||_{\infty}\le y_{j}, \sigma_{jj} = y_{j}, \forall j \in [d], \sum_{j \in [d]} y_{j}=1}, \\
	&{\bm x= \bm \mu_{\ell 1} +\bm \mu_{\ell 2}, ||\bm \mu_{\ell 1}||_{\infty}\le \alpha_{\ell}, ||\bm \mu_{\ell 2}||_{\infty}\le 1- \alpha_\ell, \forall \ell\in [m]}, \\
	&w=w_U - (w_U-w_L)\bigg(\sum_{i\in [m]} 2^{-i}\alpha_i\bigg), \\
	&{\bigg|\bigg|\sum_{i \in [n]}\bm{c}_i \bm{c}_i^{\top}\bm\delta_{i1} - w_U \bm{x} + (w_U-w_L)\sum_{\ell \in [m]} 2^{-\ell}\bm \mu_{\ell1}\bigg|\bigg|_{\infty} \le \epsilon }, \\
	& \bm \alpha \in \{0,1\}^m, \bm y \in \{0,1\}^d,
	\end{aligned}
	\end{equation}
	where $w_L, w_U$ separately denote the lower and upper bounds of SPCA, $m:= \lceil \log_2 ((w_U-w_L) \epsilon^{-1})\rceil $ and the infinite norm inequality constraints can be easily linearized. 
\end{restatable}
\begin{proof} 
	See Appendix \ref{proof_them_milp}. \qed
\end{proof}

For the proposed MILP formulation \eqref{spca_milp}, we remark that
\begin{enumerate}[(i)]
	\item This is the first-known MILP representation with arbitrary accuracy $O(\epsilon)$ in literature of SPCA; 
	\item 
	The MILP formulation \eqref{spca_milp}, although compact, involves $O(n+d+\log \epsilon^{-1})$ binary variables, $O(nd+d \log \epsilon^{-1})$ continuous variables, and $O(nd+n \log \epsilon^{-1})$ linear constraints; 
	\item In SPCA \eqref{spcacom3_eigen}, one might be curious about the choice of infinite norm. Unfortunately, as far as we are concerned, this is the only norm that leads to a compact MILP formulation; 
	\item In the MILP formulation \eqref{spca_milp}, one might consider replacing the infinite norm in the constraint ${||\sum_{i \in [n]}\bm{c}_i \bm{c}_i^{\top}\bm\delta_{i1} - w_U \bm{x} + (w_U-w_L)\sum_{i\in [m]} 2^{-i}\bm \mu_{i1}||_{\infty} \le \epsilon }$ by other norms, which will lead to different formulations (either MILP or mixed-integer conic program) and slightly different approximation bounds;
	\item Strong lower and upper bounds of SPCA $w_L,w_U$ can speed up the solution procedure;
	and 
	\item Instead of building a relatively large-scale MILP formulation \eqref{spca_milp}, one might solve $d$ number of smaller-scale MILPs by enumerating each set of a disjunction $\cup_{j\in [d]}\big\{\bm{x}: x_j=1,\|\bm{x}\|_{\infty}\leq 1 \big\}.$ 
\end{enumerate} 

The last remark is summarized in the following corollary.
\begin{corollary}
	\label{them_milp_cor}
	Given a threshold $\epsilon>0$, 
the optimal value of MILP \eqref{spca_milp} is equal to $\hat w(\epsilon)=\max_{j\in [d]}\hat w_j(\epsilon)$, where for each $j \in [d]$, $\hat w_j(\epsilon)$ is defined as
	\begin{equation}\label{spca_milp_j}
	\begin{aligned} 
	\hat{w}_j(\epsilon) := &\max_{\begin{subarray}{c}
		w, \bm{z} \in Z, \bm y, \bm{\alpha},\bm{x}, \bm \delta,\bm \mu
		\end{subarray}} w\\
	\text{\rm s.t.}\quad
	& {\bm{x} = \bm \delta_{i1}+\bm \delta_{i2}, ||\bm \delta_{i1}||_{\infty}\le z_i, ||\bm \delta_{i2}||_{\infty}\le 1- z_i, \forall i \in [n]}, \\
	&||\bm x||_{\infty}\le 1, x_j=1, \\
	&{\bm x= \bm \mu_{\ell 1} +\bm \mu_{\ell 2}, ||\bm \mu_{\ell 1}||_{\infty}\le \alpha_{\ell}, ||\bm \mu_{\ell 2}||_{\infty}\le 1- \alpha_\ell, \forall \ell\in [m]}, \\
	&w=w_U - (w_U-w_L)\bigg(\sum_{i\in [m]} 2^{-i}\alpha_i\bigg), \\
	&{\bigg|\bigg|\sum_{i \in [n]}\bm{c}_i \bm{c}_i^{\top}\bm\delta_{i1} - w_U \bm{x} + (w_U-w_L)\sum_{i\in [m]} 2^{-i}\bm \mu_{i1}\bigg|\bigg|_{\infty} \le \epsilon }, \\
	& \bm \alpha \in \{0,1\}^m,
	\end{aligned}
	\end{equation}
		where $w_L, w_U$ separately denote the lower and upper bounds of SPCA, $m:= \lceil \log_2 ((w_U-w_L) \epsilon^{-1})\rceil $ and the infinite norm inequality constraints can be easily linearized. 
\end{corollary}
Albeit being smaller-size, some MILPs defined in \Cref{them_milp_cor} might be infeasible. Since the optimal value of an infeasible maximization problem is $-\infty$ by default, the result in \Cref{them_milp_cor} still holds. However, one might need to be cautious when using this result and be aware of infeasibilities.

\subsection{Theoretical Optimality Gap}
Similar to other two exact formulations, we are also interested in deriving theoretical approximation bound for MILP formulation \eqref{spca_milp} by relaxing binary variables $\bm{z}$. Particularly, we assume that other binary variables $\bm{y},\bm{\alpha}$ can be enumerated effectively. Our results show that the theoretical optimality gap is, in general, worse than the other two bounds.
\begin{restatable}{theorem}{approxmilp}\label{thm_approx_milp}
	Given a threshold $\epsilon>0$, by enforcing the binary variables $\bm z$ to be continuous, let $\overline{w}_5(\epsilon)$ denote the optimal value of the relaxed MILP formulation \eqref{spca_milp}. Then we have $$\overline{w}_5(\epsilon)\leq \min \big\{ k(\sqrt{d}/2+1/2),\ n/k \sqrt{d} + (n-k)(\sqrt{d}/2+1/2)\big\} w^* +\epsilon\sqrt{d}.$$ 
\end{restatable}
\begin{proof}
	See Appendix \ref{proof_approx_milp}.
	\qed
\end{proof}

\section{Approximation Algorithms}\label{sec:approx_alg}
In this section, motivated by the equivalent combinatorial formulation \eqref{spcacom2}, we prove and demonstrate the tightness of the approximation ratios of the well-known greedy and local search algorithms for solving SPCA.

\subsection{Greedy Algorithm}\label{sec:greedy_alg}
The greedy algorithm has been widely used in many combinatorial problems with the cardinality constraint. The greedy algorithm in this subsection is particularly based on the combinatorial formulation \eqref{spcacom2}, which proceeds as follows: 
Given a subset $\hat S_G\subseteq [n]$ denoting the selected vectors, it aims to find a new vector from $\{\bm c_i\}_{i\in [n]\setminus \hat S_G}$ to maximize the largest eigenvalue of the sum of rank-one matrices obtained so far including the new one. The detailed implementation can be found in \Cref{alg:greedy}.

\begin{algorithm}[htbp]
	\caption{Greedy Algorithm for SPCA \eqref{spcacom2}}
	\label{alg:greedy}
	\begin{algorithmic}[1]
		\State \textbf{Input:} $n\times n$ matrix $\bm{A} \succeq 0$ of rank $d$ and integer $k \in [n]$
		\State Let $\bm{A}=\bm{C}^{\top}\bm{C}$ denote its Cholesky factorization where $\bm{C} \in \Re^{d\times n}$
		\State Let $\bm{c}_i \in \Re^d$ denote the $i$-th column vector of matrix $\bm{C}$ for each $i \in [n]$
		
		\State Let $\hat{S}_G := \emptyset$ denote the chosen set
		\For{$\ell = 1, \cdots, k$}
		\State Compute 
		$j^* \in \argmax_{j \in[n]\setminus \hat{S}}\{\lambda_{\max}(\sum_{i \in \hat{S}_G \cup\{j\}}\bm c_i \bm c_i^\top)\}$	
		\State Add $j^*$ to the set $\hat{S}_G$
		
		\EndFor

		\State \textbf{Output:} $\hat S_G$
	\end{algorithmic}
\end{algorithm}


%
The following result show that the greedy \Cref{alg:greedy} yields $1/k$-approximation ratio.
\begin{theorem}\label{prop_greedy_alg} 
The greedy \Cref{alg:greedy} yields a $k^{-1}$-approximation ratio for SPCA \eqref{spcacom2}, i.e., the output $\hat S_G$ of \Cref{alg:greedy} satisfies
\[\lambda_{\max}\bigg(\sum_{i\in \hat S_G}\bm c_i\bm c_i^\top \bigg) \ge \frac{1}{k} w^*.\]
\end{theorem}
\begin{proof}
Suppose that the optimal set of SPCA \eqref{spcacom2} is $S^*$, then we have
\[\lambda_{\max}\bigg(\sum_{i\in S^*}\bm c_i\bm c_i^\top\bigg)\leq \sum_{i\in S^*}\lambda_{\max}(\bm c_i\bm c_i^\top) \le k \max_{i\in [n]}\lambda_{\max}(\bm c_i\bm c_i^\top) \leq k\lambda_{\max}\bigg(\sum_{i\in \hat S_G}\bm c_i\bm c_i^\top \bigg),\]
where the first inequality results from the convexity of largest eigenvalue function and the last one is because at the first iteration, the greedy \Cref{alg:greedy} must choose the largest-length vector.
\qed
\end{proof}
The approximation ratio $k^{-1}$ of greedy \Cref{alg:greedy} is tight, since there exists an example whose greedy optimum is no better than $k^{-1}$. This example is presented as below.
\begin{example} \label{example}
For any integer $k\in [d]$, let $d=k+1$, $n=2k$, and the vectors $\{\bm c_i \}_{i\in [n]}\subseteq \Re^d$ be
\[\bm c_i = \begin{cases}
\bm e_i, & \text{\rm if } i \in [k], \\
 \bm e_{k+1}, & \text{\rm if } i \in [k+1,n],
\end{cases}\forall i\in [n]. \]
\end{example}

\begin{proposition} \label{prop:greed_exam}
In \Cref{example}, the output value of greedy \Cref{alg:greedy} is $k^{-1}$-away from the true optimal value of SPCA. That is, approximation ratio $k^{-1}$ of greedy \Cref{alg:greedy} is tight.
\end{proposition}
\begin{proof}
In \Cref{example}, according to the greedy \Cref{alg:greedy}, it will select $\bm c_1, \bm c_2, \cdots, \bm c_k$ at each iteration, i.e., the output set is $\hat S_G=[k]$. Thus, the resulting largest eigenvalue of greedy \Cref{alg:greedy} is equal to 1.

Apparently, the true optimal value of \Cref{example} is equal to 
\[\lambda_{\max}\bigg(\sum_{i\in [k+1, n]} \bm c_i \bm c_i^{\top}\bigg) =\lambda_{\max}\left(k \bm e_{k+1} \bm e_{k+1}^{\top}\right) =k.\]
This completes the proof.
\qed
\end{proof}

\subsection{Local Search Algorithm}\label{sec:LS_alg}
The local search algorithm can improve the existing solutions and has been successfully used to solve many interesting machine learning and data analytics problems, such as experimental design \cite{madan2019combinatorial} and maximum entropy sampling \cite{li2020best}. This subsection investigates the local search algorithm for SPCA \eqref{spcacom2} and proves its approximation ratio. 

In the local search algorithm, we start with a size-$k$ subset, and in each iteration, swap an element of chosen set with one of the unchosen set as long as it improves the largest eigenvalue. The detailed implementation can be found in \Cref{alg:localsearch}. 

%

\begin{algorithm}[htbp]
	\caption{Local Search Algorithm for SPCA \eqref{spcacom2}}
	\label{alg:localsearch}
	\begin{algorithmic}[1]
		\State \textbf{Input:} $n\times n$ matrix $\bm{A} \succeq 0$ of rank $d$ and integer $k \in [n]$
		\State Let $\bm{A}=\bm{C}^{\top}\bm{C}$ denote its Cholesky factorization where $\bm{C} \in \Re^{d\times n}$
		\State Let $\bm{c}_i \in \Re^d$ denote the $i$-th column vector of matrix $\bm{C}$ for each $i \in [n]$
		\State Initialize a size-$k$ subset $\hat{S}_L \subseteq [n]$ 
		\Do
		\For{each pair {$(i,j) \in \hat{S}_L \times ([n]\setminus \hat{S}_L)$}}
		\If{$\lambda_{\max} \left(\sum_{\ell \in \hat{S}_L \cup \{j\} \setminus \{i\}} \bm{c}_\ell \bm{c}_\ell^{\top}\right) > \lambda_{\max} \left(\sum_{\ell \in \hat{S}_L} \bm{c}_\ell \bm{c}_\ell^{\top}\right)$}
		\State Update $\hat{S}_L := \hat{S}_L \cup \{j\} \setminus \{i\}$
		\EndIf
		\EndFor
		\doWhile{there is still an improvement}
		\State \textbf{Output:} $\hat S_L$
	\end{algorithmic}
\end{algorithm}
\begin{theorem}\label{prop_LS_alg} 
The local search \Cref{alg:localsearch} returns a $k^{-1}$-approximation ratio of SPCA, i.e., the output $\hat S_{L}$ of the local search \Cref{alg:localsearch} satisfies
\[\lambda_{\max}\bigg(\sum_{i\in \hat S_{L}}\bm c_i\bm c_i^\top\bigg)\ge \frac{1}{k} w^*.\]
\end{theorem}
\begin{proof}
First, 
for each $j\in [n]$, we will show that
\begin{align} \label{eq:lsineq1}
\lambda_{\max}\bigg(\sum_{\ell \in \hat S_{L}} \bm c_{\ell} \bm c_{\ell}^{\top} \bigg) \ge \lambda_{\max}(\bm c_j \bm c_j^{\top}).
\end{align}
To prove it, there are two cases to be discussed: whether $j$ belongs to $\hat S_L$ or not. The monotonicity of the largest eigenvalue of sum of positive semi-definite matrices implies that the inequality \eqref{eq:lsineq1} holds if $j\in \hat S_L$. If $j \in [n]\setminus \hat S_L$, then the local optimality condition implies that there exist $i \in \hat S_L$ such that
\[ \lambda_{\max} \bigg(\sum_{\ell \in \hat{S}_L} \bm{c}_\ell \bm{c}_\ell^{\top}\bigg) \ge \lambda_{\max} \bigg(\sum_{\ell \in \hat{S}_L \cup \{j\} \setminus \{i\}} \bm{c}_\ell \bm{c}_\ell^{\top}\bigg) \geq \lambda_{\max}(\bm c_j \bm c_j^{\top}),\]
where the second inequality is due to the monotonicity of the largest eigenvalue of sum of positive semi-definite matrices.
%
%

Second, suppose $S^*$ to be the optimal solution to SPCA \eqref{spcacom2}, by inequality \eqref{eq:lsineq1}, then we have
\begin{align*}
w^* = \lambda_{\max} \bigg(\sum_{i \in {S^*}} \bm{c}_i \bm{c}_i^{\top}\bigg) \le \sum_{i \in S^* } \lambda_{\max}(\bm c_i \bm c_i^{\top}) \le k \lambda_{\max} \bigg(\sum_{\ell \in \hat{S}_L} \bm{c}_{\ell} \bm{c}_{\ell}^{\top}\bigg),
\end{align*}
where the first inequality is because of the convexity of function $\lambda_{\max}(\cdot)$.
\qed
\end{proof}

We remark that \Cref{example} also confirms the tightness of our analysis for local search \Cref{alg:localsearch}.

\begin{proposition} \label{prop_LS}
In \Cref{example},	the output value of local search \Cref{alg:localsearch} is $k^{-1}$-away from optimal value of SPCA. That is, approximation ratio $k^{-1}$ of local search \Cref{alg:localsearch} is tight.
\end{proposition}
\begin{proof}
In \Cref{example}, we show that the initial subset $\hat S_L=[k]$ already satisfies the local optimality condition.

Indeed, for each pair $(i,j) \in \hat S_L \times ([n]\setminus \hat S_L)$, we have
\[ \lambda_{\max} \bigg(\sum_{\ell \in \hat{S}_L \cup \{j\} \setminus \{i\}} \bm{c}_\ell \bm{c}_\ell^{\top}\bigg) = \lambda_{\max} (\bm I_d - \bm e_i \bm e_i^{\top}) =1= \lambda_{\max} (\bm I_d - \bm e_{d} \bm e_{d}^{\top}) = \lambda_{\max} \bigg(\sum_{\ell \in \hat{S}_L} \bm{c}_\ell \bm{c}_\ell^{\top}\bigg), \]
where the identities follow the construction of $\{\bm c_i\}_{i\in [n]}$ in \Cref{example}.

Therefore, the set $\hat S_L$ achieves the local optimum with largest eigenvalue of $1$. Since the optimal value of SPCA is $w^*=k$, the approximation ratio of set $\hat S_L$ is equal to $k^{-1}$.
%
	\qed
\end{proof}

As an improved heuristic, local search \Cref{alg:localsearch} can use the output of the greedy \Cref{alg:greedy} as an initial solution. The results in \Cref{prop_LS_alg} and \Cref{prop_LS} imply that the integrated algorithm still yields a $k^{-1}$-approximation ratio of SPCA, while for solving the practical instances, our numerical study shows that the integrated algorithm in fact works very well. 
Since the greedy \Cref{alg:greedy} and local search \Cref{alg:localsearch} repeatedly require to compute the largest eigenvalues, at each iteration, we can apply the power iteration method to efficiently calculate the largest eigenvalues \cite{semlyen1995efficient} and use the eigenvectors from the previous iterations as a warm-start.

Finally, we remark that there is only one swap in the local search \Cref{alg:localsearch}. We can improve it by increasing the number of swapping elements at each iteration, termed \textit{$s$-swap local search} with $s\in [k]$. The following result shows that $s$-swap local search can indeed achieve a better approximation ratio. 
\begin{corollary}
The approximation ratio of $s$-swap local search is $sk^{-1}$ for any $s\in [k]$. The approximation ratio is tight.
\end{corollary}
\begin{proof}
First, let set $ \hat S_{L}$ denote the indices of selected vectors by $s$-swap local search algorithm. Then following the same proof as that in \Cref{prop_LS_alg}, 
for any size-$s$ set $T \subseteq [n]$, we have
\begin{align} \label{ineq_sswap}
\lambda_{\max}\bigg(\sum_{i \in \hat S_{L}} \bm c_{i} \bm c_{i}^{\top} \bigg) \ge \lambda_{\max}\bigg(\sum_{i \in T} \bm c_{i} \bm c_{i}^{\top} \bigg).
\end{align}

Let $S^*$ denote the optimal solution to SPCA \eqref{spcacom2}, using the result \eqref{ineq_sswap}, the optimal value of SPCA $w^*$ is upper bounded by
\begin{align*}
w^* = \lambda_{\max} \bigg(\sum_{i \in {S^*}} \bm{c}_i \bm{c}_i^{\top}\bigg) = \lambda_{\max} \bigg(\frac{1}{\binom{k-1}{s-1}} \sum_{T \subseteq {S^*}, |T|=s} \sum_{i\in T} \bm{c}_i \bm{c}_i^{\top}\bigg) \le \frac{\binom{k}{s}}{\binom{k-1}{s-1}} \lambda_{\max} \bigg(\sum_{i \in \hat{S}_L} \bm{c}_{i} \bm{c}_{i}^{\top}\bigg) = \frac{k}{s}\bigg(\sum_{i \in \hat{S}_L} \bm{c}_{i} \bm{c}_{i}^{\top}\bigg).
\end{align*}

Second, to show the tightness, let us consider the following example.
\begin{example}\label{example2}
For any integer $k\in [d]$, let $d=k+1$, $n=(s+1)k$, and the vectors $\{\bm c_i \}_{i\in [n]}\subseteq \Re^d$ be
\[\bm c_i = \begin{cases}
\bm e_i, & \text{\rm if } i \in [k], \\
\vdots\\
\bm e_{i-(s-1)k}, & \text{\rm if } i \in [(s-1)k+1,sk], \\
 \bm e_{k+1}, & \text{\rm if } i \in [sk+1,n],
\end{cases}\forall i\in [n]. \]
\end{example}
In \Cref{example2}, we show that the subset $\hat S_L=[k-s+1]\cup\{\ell k+1\}_{\ell\in [s-1]}$ satisfies the $s$-swap local optimality condition.

Indeed, for each pair $(T_1,T_2)$ such that $T_1\subseteq\hat S_L, T_2\subseteq ([n]\setminus \hat S_L)$ with $|T_1|=|T_2|=s$, we have
\[ \lambda_{\max} \bigg(\sum_{\ell \in \hat{S}_L \cup T_2 \setminus T_1} \bm{c}_\ell \bm{c}_\ell^{\top}\bigg) \leq s. \]
Therefore, the set $\hat S_L$ achieves $s$-swap local optimum with largest eigenvalue of $s$. Since the optimal value of SPCA is $w^*=k$, the approximation ratio of set $\hat S_L$ is equal to $sk^{-1}$ for SPCA.
\qed
\end{proof}
Albeit theoretically sound, $s$-swap local search with $s\geq 2$ might not be practical since it involves $O(n^2)$ swaps at each iteration. Therefore, in the numerical study, we use the simple local search \Cref{alg:localsearch}, which already works very well.

\section{Numerical Study}\label{sec:computation}
In this section, we conduct numerical experiments on six datasets with number of features $n$ ranging from 13 to 2365 to demonstrate the computational efficiency and the solution quality of the MISDP \eqref{sdp_two_eq}, MISDP \eqref{sdp_one_stronger}, and MILP \eqref{spca_milp} for exactly solving SPCA, the continuous relaxations \eqref{sdp_two_eq_re}, \eqref{sdp_one_stronger_rel} and heuristic Algorithms \ref{alg:greedy}, \ref{alg:localsearch} for approximately solving SPCA. All the methods in this section are coded in Python 3.6 with calls to Gurobi 9.0 and MOSEK 9.0 on a personal PC with 2.3 GHz Intel Core i5 processor and 8G of memory. The codes and data are available at \url{https://github.com/yongchunli-13/Sparse-PCA}.

\subsection{\textit{Pitprops} Dataset} 

We first test the proposed three exact SPCA formulations \eqref{sdp_two_eq}, \eqref{sdp_one_stronger}, \eqref{spca_milp} and their continuous relaxations to solve a commonly-used benchmark instance, 
\textit{Pitprops} dataset \citet{jeffers1967two}, which consists of 13 features (i.e., $n=13$). In this instance, the computational results of seven different cases with $k$ chosen from $\{4,\cdots, 10\}$ are displayed in \Cref{table13-exact}, \Cref{table13-relaxation}, and \Cref{table13_lower}. 

For each testing case, we solve two MISDP formulations \eqref{sdp_two_eq} and \eqref{sdp_one_stronger} using the branch and cut method. As for the MILP \eqref{spca_milp}, it can be simply solved in Gurobi.
Throughout the numerical study of MILP \eqref{spca_milp}, we set $\epsilon = 10^{-4}$, use the best SDP relaxation values as the upper bound $w_U$, and use the local search \Cref{alg:localsearch} to compute the lower bound $w_{L}$. 
As the newly released Gurobi 9.0 is able to solve the non-convex quadratic program, thus for the purpose of comparison, we further use Gurobi to solve the following SPCA formulation
\begin{align}\label{Gurobi}
 w^*:= \max_{\bm z\in Z, \bm x\in \Re^n} \left\{\bm x^{\top} \bm A\bm x: ||\bm x||_2=1, ||\bm x||_1\le \sqrt{k}, |x_i|\le z_i, \forall i \in [n]\right\}.
\end{align}

The computation results of the exact methods are shown in \Cref{table13-exact}. In particular, we let \textbf{time(s)} denote the running time in seconds of each case and let \textbf{Gurobi} denote the performance of Gurobi for solving SPCA \eqref{Gurobi}.
In \cref{table13-exact}, we see that all the SPCA formulations \eqref{sdp_two_eq}, \eqref{sdp_one_stronger}, \eqref{spca_milp} can be solved to optimality within seconds, which demonstrates the efficiency of the proposed formulations. We also compare the numerical performance of the MILP formulation \eqref{spca_milp} with formulation \eqref{Gurobi} using the Gurobi solver, and it is clear that MILP is more efficient and stable. Especially for the case of $k=10$, Gurobi has trouble finding the optimal solution of SPCA \eqref{Gurobi}. 

\begin{table}[!h] \centering
\caption{Computational results of exact values with \textit{Pitprops} dataset}
\setlength{\tabcolsep}{3pt}\renewcommand{\arraystretch}{1.2}
\begin{tabular}{c |r |r r |r r | r r | r r }
\hline 
\multicolumn{1}{c|}{$n$=13} & \multicolumn{1}{c|}{SPCA} & \multicolumn{2}{c|}{MISDP \eqref{sdp_two_eq}} & \multicolumn{2}{c|}{MISDP \eqref{sdp_one_stronger} } & \multicolumn{2}{c|}{MILP \eqref{spca_milp}} & \multicolumn{2}{c}{Gurobi }
\\ \cline{1-10} 
$k$& \multicolumn{1}{c|}{$w^*$} & \multicolumn{1}{c}{$w^*$} & \multicolumn{1}{c|}{time(s)} & \multicolumn{1}{c}{$w^*$} & \multicolumn{1}{c|}{time(s)} & \multicolumn{1}{c}{$\hat{w}(\epsilon)$}& \multicolumn{1}{c|}{time(s)}&
\multicolumn{1}{c}{$w^*$}& \multicolumn{1}{c}{time(s)}
\\
\hline
4 & 2.9375 & 2.9375 & 1 &2.9375 &2 &2.9375&1&2.9375&1 \\ 
5 & 3.4062 &3.4062&1 &3.4062 &2& 3.4062&1&3.4062&1\\
6 & 3.7710& 3.7710& 1 &3.7710 &2&3.7710&2&3.7710&1 \\
7 & 3.9962&3.9962&1 &3.9962 &1&3.9962&1&3.9962&3 \\
8 & 4.0686&4.0686&1 &4.0686 &2&4.0686&2&4.0686&12\\
9& 4.1386& 4.1386&1 &4.1386 &2&4.1386&1&4.1387&30\\
10& 4.1726&4.1726 &1 &4.1726 &1&4.1726&1&4.1441&83\\
\hline
\end{tabular}%
\label{table13-exact}
\end{table}

Although the theoretical optimality gaps of the proposed SDP relaxations \eqref{sdp_two_eq_re} and \eqref{sdp_one_stronger_rel} are the same, these gaps in practice can be much smaller and can be significantly different from each other. We use MOSEK to solve both SDP relaxations. The numerical results can be found in \Cref{table13-relaxation}, where the SDP relaxation \eqref{eq_sdp_d2005direct} proposed by \citet{d2005direct} is presented as a benchmark comparison. In \Cref{table13-relaxation}, we use \textbf{gap(\%)} to denote the optimality gap, which is computed as
$100 \times {(\textrm{Upper Bound} - w^*)}/{w^*}$. 
It can be seen that the second SDP relaxation \eqref{sdp_one_stronger_rel} is superior to the first SDP relaxation \eqref{sdp_two_eq_re} on the first five cases. When $k$ is close to $n$, the first SDP relaxation \eqref{sdp_two_eq_re} can be better. This finding is consistent with remarks after \Cref{thm_model2_relax}. In addition, as proved in \Cref{prop_bound_comparison_model1}, we see that the second SDP relaxation \eqref{sdp_one_stronger_rel} always outperforms the bound \eqref{eq_sdp_d2005direct} by \citet{d2005direct}. Finally, the second SDP relaxation \eqref{sdp_one_stronger_rel} and the bound \eqref{eq_sdp_d2005direct} by \citet{d2005direct} are also not comparable.

\begin{table}[!h]
\caption{Computational results of upper bounds with \textit{Pitprops} dataset}
\centering
\setlength{\tabcolsep}{2pt}\renewcommand{\arraystretch}{1.2}
\begin{tabular}{c |r |r r |r r r | r r r }
\hline 
\multicolumn{1}{c|}{$n$=13} & \multicolumn{1}{c|}{SPCA} & \multicolumn{2}{c|}{Benchmark \eqref{eq_sdp_d2005direct}} & \multicolumn{3}{c|}{SDP Relaxation \eqref{sdp_two_eq_re}} & \multicolumn{3}{c}{SDP Relaxation \eqref{sdp_one_stronger_rel}}
\\ \cline{1-10} 
$k$& \multicolumn{1}{c|}{$w^*$} & \multicolumn{1}{c}{$\overline{w}_4$} &\multicolumn{1}{r|}{gap($\%$)} & \multicolumn{1}{c}{$\overline{w}_1$} &\multicolumn{1}{c}{gap($\%$)} & \multicolumn{1}{c|}{time(s)} & \multicolumn{1}{c}{$\overline{w}_3$} &\multicolumn{1}{c}{gap($\%$)} & \multicolumn{1}{c}{time(s)} \\
\hline
4 & 2.9375 & 3.0172 & 2.71 &3.1065&5.75 & 0.51 &2.9495 &0.41 &0.13 \\ 
5 & 3.4062 &3.4581&1.52 &3.4868& 2.37 &0.55 &3.4124 &0.18 & 0.18\\
6 & 3.7710&3.8137& 1.13 &3.7859 & 0.39 & 0.52 &3.7767 &0.15 &0.15\\
7 & 3.9962&4.0316&0.89 &3.9962 &0.00 &0.43 &3.9962 &0.00 &0.15\\
8 & 4.0686&4.1448&1.87 &4.0805&0.29 &0.29 &4.0793 &0.26 &0.17\\
9& 4.1386&4.2063&1.64 &4.1386 &0.00 & 0.00 &4.1398 &0.03 &0.15\\
10& 4.1726&4.2186&1.10 &4.1763&0.09 & 0.09 &4.1778 &0.12 &0.16\\
\hline
\end{tabular}%
\label{table13-relaxation}
\end{table}

\Cref{table13_lower} presents the objective values and optimality gaps of the proposed approximation algorithms for solving the \textit{Pitprops} instance, where we let \textbf{LB} denote the lower bound and compute \textbf{gap(\%)} by $100\times (w^*-\text{LB})/w^*$. Note that we  initialize the local search \Cref{alg:localsearch} by the output of greedy \Cref{alg:greedy}. To further improve the two algorithms, at each iteration, we employ the power iteration method to efficiently compute the largest eigenvalues \cite{semlyen1995efficient} and warm-start it with the good-quality eigenvectors from the previous iterations. In \Cref{table13_lower}, we see that greedy \Cref{alg:greedy} and local search \Cref{alg:localsearch} successfully find the optimal solutions and outperforms the truncation algorithm proposed by \cite{chan2016approximability}.
\begin{table}[!h]
\caption{Computational results of lower bounds with \textit{Pitprops} dataset}
\centering
\setlength{\tabcolsep}{2pt}\renewcommand{\arraystretch}{1.2}
\begin{tabular}{c |r |r r r |r r r | r r r }
\hline 
\multicolumn{1}{c|}{$n$=13} & \multicolumn{1}{c|}{SPCA} & \multicolumn{3}{c|}{Truncation algorithm \cite{chan2016approximability}} & \multicolumn{3}{c|}{Greedy \Cref{alg:greedy}} & \multicolumn{3}{c}{Local Search \Cref{alg:localsearch}}
\\ \cline{1-11} 
$k$& \multicolumn{1}{c|}{$w^*$} & \multicolumn{1}{c}{LB} &\multicolumn{1}{c}{gap($\%$)} &
\multicolumn{1}{c|}{time(s)} &
\multicolumn{1}{c}{LB} &\multicolumn{1}{c}{gap($\%$)} & \multicolumn{1}{c|}{time(s)} & \multicolumn{1}{c}{LB} &\multicolumn{1}{c}{gap($\%$)} & \multicolumn{1}{c}{time(s)} \\
\hline
4 & 2.9375 & 2.8913 & 1.57 &1e-3&2.9375 & 0.00 &1e-3 &2.9375 &0.00 &1e-2 \\ 
5 & 3.4062 & 3.3951 & 0.32 &1e-3& 3.4062 &0.00 &1e-3 &3.4062 & 0.00 &1e-2\\
6 & 3.7710 & 3.7576& 0.36 &1e-3 & 3.7710 & 0.00 &1e-2 &3.7710 &0.00 & 1e-2\\
7 & 3.9962 & 3.9929 & 0.08 &1e-3 &3.9962 &0.00 &1e-2 &3.9962 &0.00 & 1e-2\\
8 & 4.0686 & 4.0648& 0.09 &1e-3&4.0686 &0.00 &1e-2 &4.0686 &0.00 & 1e-2\\
9& 4.1386 & 4.1313& 0.18 &1e-3 &4.1386 & 0.00 &1e-2 &4.1386 &0.00 & 1e-2\\
10& 4.1726 & 4.0094& 3.91 &1e-3&4.1726 & 0.00 &1e-2 &4.1726 &0.00 & 1e-2\\
\hline
\end{tabular}%
\label{table13_lower}
\end{table}

\subsection{Four Larger-scale Datasets}
In this subsection, we conduct experiments on four larger instances from \citet{dey2018convex} to further testify the efficiency of our proposed methods for SPCA, which are \textit{Eisen-1}, \textit{Eisen-2}, \textit{Colon} and \textit{Reddit} with $n=$79, 118, 500, and 2000. Since the MILP formulation \eqref{spca_milp} consistently outperforms two MISDP formulations \eqref{sdp_two_eq} and \eqref{sdp_one_stronger}. Thus, in this set of numerical experiments, we will stick to the MILP formulation \eqref{spca_milp}.

We first compare the performances of different heuristic methods using the \textit{Reddit} dataset with $n=2000$ and $k\in \{10,\ldots,70\}$. Thus, there are 7 cases in total. We implement the greedy \Cref{alg:greedy} and the local search \Cref{alg:localsearch} and compare them with the best-known truncation algorithm proposed by \cite{chan2016approximability}. The numerical results are shown in \Cref{table_large_local}. We see that the local search \Cref{alg:localsearch} provides the highest-quality solution of the three. The greedy \Cref{alg:greedy} is almost equally as good as the truncation algorithm. Although the local search \Cref{alg:localsearch} takes the longest running time, the running time is quite reasonable given the size of the testing cases. Hence, our computation experiments show that the local search \Cref{alg:localsearch} consistently outperforms the other two methods within a reasonably short time. Thus, we recommend using this algorithm to solve practical problems.

\begin{table}[!h]
	\caption{Computational results of lower bounds with \textit{Reddit} dataset}
	\centering
	\setlength{\tabcolsep}{2pt}\renewcommand{\arraystretch}{1.2}
	\begin{tabular}{c |r r |r r | r r }
		\hline 
		\multirow{2}{3.5em}{$n$=2000} & \multicolumn{2}{c|}{Truncation } &\multicolumn{2}{c|}{ Greedy} & \multicolumn{2}{c}{Local Search }\\
		& \multicolumn{2}{c|}{algorithm \cite{chan2016approximability}} & \multicolumn{2}{c|}{ \Cref{alg:greedy}} &\multicolumn{2}{c}{ \Cref{alg:localsearch}} 
		\\ \cline{1-7} 
		$k$& \multicolumn{1}{c}{LB} & \multicolumn{1}{c|}{time (s)} & \multicolumn{1}{c}{LB} & \multicolumn{1}{c|}{time (s)} & \multicolumn{1}{c}{LB} & \multicolumn{1}{c}{time (s)}\\
		\hline
		10 & 1482.3205 & 3 & 1521.3081 & 1 &1521.3083 & 9 \\ 
		20 & 1666.2397	&2	&1670.4712	&4	&1684.3943	&59
\\
		30 & 1953.3711	&2	&1856.2875	&7	&1953.7502	&92
\\
		40 & 2203.1715	&2	&2123.5635	&10	&2208.2452	&208
\\
		50 & 2311.2407	&2	 &2289.0371	&13	&2322.8204	&207
\\
		60 & 2427.2685	&3 &2402.8345	&16 &2441.7020	&202
\\
		70 & 2475.9581	&2	&2488.8991	&19 &2494.6142	&193\\
		\hline
	\end{tabular}%
	\label{table_large_local}
\end{table}

Next, we obtain the local search \Cref{alg:localsearch}, the continuous relaxation bounds and exact values of SPCA on the four instances, i.e., \textit{Eisen-1}, \textit{Eisen-2}, \textit{Colon} and \textit{Reddit}. For these instances, MOSEK fails to solve our proposed SDP relaxations \eqref{sdp_two_eq_re} and \eqref{sdp_one_stronger_rel}. Thus, instead, we use the subgradient method to solve the continuous relaxation formulations \eqref{eq_overline_w_2} and \eqref{eq_overline_w_4}. 
For the MILP formulation \eqref{spca_milp}, we set the time limit of Gurobi to be an hour. The computational results are presented in \Cref{table_large}, where we let \textbf{UB} denote the upper bound of SPCA, let \textbf{VAL} denote the best lower bound of MILP \eqref{spca_milp} found if the time limit is reached, and let \textbf{MIPgap($\%$)} denote the percentage of output MIP Gap from Gurobi.
For these instances, we see that the local search \Cref{alg:localsearch} still performs very well and the subgradient method is also efficient to solve the continuous relaxation \eqref{eq_overline_w_2}. The continuous relaxation \eqref{eq_overline_w_4} turns out to be very difficult to compute, and even more difficult than the MILP formulation \eqref{spca_milp}. For the instance \textit{Eisen-1}, we see that both the MILP formulation \eqref{spca_milp} and local search \Cref{alg:localsearch} can find the optimal solutions. This further demonstrates the effectiveness of the local search \Cref{alg:localsearch}.

\begin{table}[!h] 
\caption{Computational results of lower bounds, upper bounds and exact values with four larger instances}
\centering
 \fontsize{10}{10}\selectfont 
\setlength{\tabcolsep}{1.5pt}\renewcommand{\arraystretch}{1.2}
\begin{tabular}{c|c c | r r| r r| r r| r r r }
\hline 
\multirow{3}{3em}{Data} & \multicolumn{2}{c|}{Case} & \multicolumn{2}{c|}{Local Search } & \multicolumn{2}{c|}{Continuous} & \multicolumn{2}{c|}{Continuous}& 
 \multicolumn{3}{c}{MILP \eqref{spca_milp}}\\
 & & & \multicolumn{2}{c|}{ \Cref{alg:localsearch}} &\multicolumn{2}{c|}{Relaxation \eqref{eq_overline_w_2} } &\multicolumn{2}{c|}{Relaxation \eqref{eq_overline_w_4}} 
\\ \cline{2-12} 
& $n$& $k$& \multicolumn{1}{c}{LB} & \multicolumn{1}{c|}{time(s)} & \multicolumn{1}{c}{UB} & \multicolumn{1}{c|}{time(s)} &
\multicolumn{1}{c}{UB} & \multicolumn{1}{c|}{time(s)} & \multicolumn{1}{c}{VAL} & \multicolumn{1}{c}{MIPgap($\%$)} & \multicolumn{1}{c}{time(s)} \\
\hline
Eisen-1 &79&10 & 17.3355& 1 & 17.9144& 14& 17.7571& 126& 17.3355 & 0.00 &34 \\ 
 &79& 20 & 17.7195 & 1 & 18.1309& 13& 18.0362& 85& 17.7195 & 0.00& 125\\
\hline 
\hline
Eisen-2 &118&10 & 11.7182 & 1 & 13.8732& 89& -&- & 11.7182 & 18.39 &3600 \\ 
&118& 20 & 19.3228 & 1 & 22.9268 & 90& -&- & 19.3228 & 18.65& 3600\\
\hline 
\hline
Colon &500&10 & 2641.2289 & 1 & 2901.1105 & 342&- &- & 2641.2289&9.84 &3600 \\ 
&500& 20 & 4255.6941 & 3 & 4833.1900& 344&- &- &4255.6941 & 13.57& 3600\\
\hline 
\hline
Reddit &2000&10 & 1521.3083 & 9 & 1867.9965 & 1198&- & -& - & - &- \\ 
&2000& 20 & 1684.3943& 59 & 2184.2436 & 1241&- &- & - & - &-\\
\hline 
\end{tabular}%
\label{table_large}
\end{table}

\subsection{\textit{Drugabuse} Dataset}
\begin{figure}[h]
\centering
		\includegraphics[width=11cm,height=8.5cm]{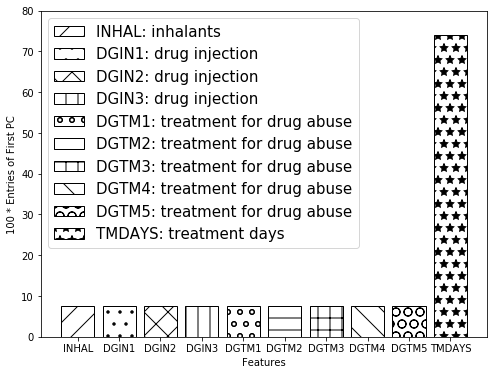}
	\caption{10 features selected by local search \Cref{alg:localsearch} for \textit{Drugabuse} dataset}
	\label{cfig}
\end{figure}

We finally apply the proposed local search \Cref{alg:localsearch} to the \textit{Drugabuse} Dataset with $n=2365$ features, where the dataset comes from a questionnaire collected by the National Survey on Drug Use and Health (NSDUH) in 2018. 
It has been reported \cite{overdose2018understanding} that with the growing illicit online sale of controlled substances, deaths attributable to opioid-related drugs have been more than quadrupled in the U.S. since 1999. 
Thus, it is important to select a handful of features that the researchers can focus on for further exploration. Indeed, SPCA is a good tool to reduce the complexity and improve the interpretability of the machine learning algorithms by selecting the most important features. Our numerical finding of the case of $k=10$ is illustrated in \Cref{cfig}, where the vertical values correspond to the selected features of the first PC, which are scaled by 100. 
We see that among 10 features, there are three categories (i.e., inhalants, drug injection, drug treatment), which are important for analyzing drug abuse. 
In particular, SPCA selects 6 features related to drug treatment, which is consistent with the literature \cite{coughlin2020machine,volkow2007dopamine} that the treatment records of drug abuse are informative and important. Three drug injection questions have been designed to understand the injection experience of different special drugs, and it is well known that drug injection users are at high risk for HIV and other blood-borne infections \cite{ompad2005childhood, thomas1995correlates}. Inhalants feature, corresponding to various accessible products that can easily cause addictions, significantly contributes to the increase of drug abuse \cite{breakey1974hallucinogenic,de1984inhalant}.

\section{Extension to the Rank-one Sparse Singular Value Decomposition (R1-SSVD)}\label{sec_svd}

In this section, we extend the proposed formulations and theoretical results to the rank-one sparse singular value decomposition (R1-SSVD). R1-SSVD has been successfully used to analyze the row-column associations within high-dimensional data (see, e.g., \cite{min2016l0,lee2010biclustering,sill2011robust}). The goal of R1-SSVD is to find the best submatrix (possibly non-square) of a particular size whose largest singular value is maximized, from a given matrix.

Formally, R1-SSVD can be formulated as
\begin{align} \label{ssvd}
\text{\rm (R1-SSVD)} \quad w_{\rm SVD}^* := \max_{\bm{u}\in \Re^{m}, \bm{v} \in\Re^n} \left\{\bm{u}^{\top}\bm{A}\bm{v}: ||\bm{u}||_{2} =1, ||\bm{v}||_{2}=1,||\bm{u}||_0 = k_1 , ||\bm{v}||_0= k_2 \right\},
\end{align}
where the matrix $\bm{A} \in \Re^{m \times n}$ is known, $m,n$, and $k_1\in [m]$ and $k_2\in [n]$ are positive integers. 

Our reduction of R1-SSVD \eqref{ssvd} to SPCA \eqref{spca} follows from the development of an augmented symmetric matrix $\overline{\bm{A}} \in \S^{m+n}$
\begin{align} \label{eqmatrixA}
\overline{\bm{A}}=
\begin{bmatrix} 
\bm 0 & \bm A \\
\bm{A}^{\top} &\bm 0 
\end{bmatrix}.
\end{align}
Let $\bm{x}:=[\bm u^{\top},
\bm v^{\top}]^\top$ denote an $(m+n)$-dimensional vector. According to the identity
\[ \bm x^{\top} \overline{\bm{A}} \bm x = \begin{bmatrix} 
\bm u^{\top} & \bm v^{\top} \\
\end{bmatrix} 
\begin{bmatrix} 
\bm 0 & \bm A \\
\bm{A}^{\top} &\bm 0 
\end{bmatrix} 
\begin{bmatrix} 
\bm u \\
\bm v 
\end{bmatrix} =
2\bm{u}^{\top}\bm{A}\bm{v}, \]
then R1-SSVD \eqref{ssvd} can be reformulated as
\begin{align} \label{eqssvd}
w_{\rm SVD}^* := \frac{1}{2}\max_{\bm{x}\in \Re^{m+n}} \left\{ \bm{x}^{\top}
\overline{\bm A}
\bm{x} : ||\bm{x}_{1:m}||_{2} =1, ||\bm{x}_{m+1:m+n}||_{2}=1,||\bm{x}_{1:m}||_0 = k_1 , ||\bm{x}_{m+1:m+n}||_0 = k_2 \right\},
\end{align}
where we let $\bm{x}_{1:m}$ denote the collection of $m$ entries of vector $\bm x$ from index set $[m]$ and $\bm{x}_{m+1, m+n}$ denote the $n$ entries of $\bm x$ from index set $[m+1, m+n]$. In R1-SSVD \eqref{eqssvd}, we enforce the sparse restrictions on both $\bm{x}_{1:m}$ and $\bm{x}_{m+1, m+n}$. Thus, the R1-SSVD \eqref{eqssvd} can be viewed as a special case of the conventional SPCA \eqref{spca}, where $\bm{A}$ is symmetric but not positive semi-definite and there are two sparsity constraints instead of one. 

Similarly, introducing binary variable $z_i=1$ if $i$th column of matrix $\overline{\bm A}$ is chosen, 0, otherwise, we can linearize the zero-norm constraints and recast R1-SSVD \eqref{eqssvd} as
\begin{align} \label{eqssvd2}
w_{\rm SVD}^* := \frac{1}{2}\max_{\bm{x}\in \Re^{m+n},\bm{z}\in Z_{\rm SVD}} &\bigg\{ \bm{x}^{\top}
\overline{\bm A}
\bm{x} : ||\bm{x}_{1:m}||_{2} =1, ||\bm{x}_{m+1:m+n}||_{2}=1,|x|_i\leq z_i, \forall i\in [m+n]\bigg\},
\end{align}
where set $Z_{\rm SVD}$ is defined as 
$$Z_{\rm SVD}:=\bigg\{\bm{z}\in \{0,1\}^{m+n}:\sum_{i\in [m]} z_i = k_1 , \sum_{i\in [m+1,m+n]} z_i = k_2 \bigg\}.$$


The following lemma inspires us three exact mixed-integer formulations for R1-SSVD \eqref{eqssvd2}.
\begin{restatable}{lemma}{lemrssvd} \label{lem:r1ssvd}
Given a matrix $\bm{A} \in \Re^{m \times n}$, consider its augmented counterpart $\overline{\bm{A}}$ defined in \eqref{eqmatrixA}, two integers $k_1 \in [m]$ and $k_2 \in [n]$, and three subsets $S, S_1,S_2\subseteq [m+n]$ such that ${{S \subseteq [m+n], |S|=k_1+k_2}}$, $S_1=S\cap [m], |S_1|=k_1$ and $S_2=S\cap [m+1,m+n], |S_2|=k_2$. Then the following identities must hold:
 \begin{enumerate}[(i)]
\item The eigenvalues of the augmented submatrix $\overline{\bm{A}}_{S,S}$ are the singular values of submatrix $\bm{A}_{S_1, S_2}$ and their negations;
\item $\sigma_{\max}(\bm A_{S_1,S_2}) = \lambda_{\max}(\overline{ \bm{A} }_{S,S}) = 1/2\max_{\bm{x}\in \Re^{k_1+k_2}} \{ \bm{x}^{\top}
\overline{\bm A}
\bm{x} : ||\bm{x}_{1:k_1}||_{2} =1, ||\bm{x}_{k_1+1:k_1+k_2}||_{2}=1 \} =1/2\max_{\bm{X} \in \S_+^{k_1+k_2}} \left \{ \tr(\overline{\bm{A}}_{S,S}\bm{X}) : \sum_{j\in [k_1]} X_{jj}=1, \sum_{i\in [k_1+1, k_1+k_2]} X_{ii}=1 \right \}$.
 \end{enumerate}

\end{restatable}
\begin{proof}
See Appendix \ref{proof_lem_r1ssvd}. \qed
\end{proof}

Notably, Part (ii) in \Cref{lem:r1ssvd} shows that R1-SSVD is equivalent to the following combinatorial optimization problem
\begin{align} \label{eqssvd22}
w_{\rm SVD}^* :=\max_{S\subseteq [m+n]} &\bigg\{ \lambda_{\max}(\overline{\bm{A}}_{S,S}): |S\cap [m]|=k_1,|S\cap [m+1,m+n]|=k_2 \bigg\}.
\end{align}

The next four subsections present MISDP formulations (I) and (II), a MILP formulation, and approximation algorithms, respectively.

\subsection{MISDP Formulation (I)}

The fact that matrix $\overline{\bm A}$ is symmetric but not positive semi-definite impedes us to directly apply the results in \Cref{sec:sdpone}. Fortunately, a simple remedy by adding a new matrix $\sigma_{\max}(\bm A)\bm{I}_{m+n}$ to $\overline{\bm A}$ fixes this issue. That is, let us define
\begin{align} \label{eq_Astar}
\overline{\bm A}^\#:=\overline{\bm A}+\sigma_{\max}(\bm A)\bm{I}_{m+n},
\end{align}
which is indeed positive semi-definite according to Part (i) in \Cref{lem:r1ssvd}. More importantly, the new matrix $\overline{\bm A}^\#$ preserves all the sparsity properties of the original one $\overline{\bm A}$.

Thus, the combinatorial optimization R1-SSVD \eqref{eqssvd2} is equivalent to
\begin{align} \label{eqssvd2_sdp1}
w_{\rm SVD}^* :=\max_{S\subseteq [m+n]} &\bigg\{ \lambda_{\max}(\overline{\bm{A}}^\#_{S,S}): |S\cap [m]=k_1,|S\cap [m+1,m+n]=k_2 \bigg\}-\sigma_{\max}(\bm A).
\end{align}

Now all the results in \Cref{sec:sdpone} are directly applicable to R1-SSVD \eqref{eqssvd2_sdp1}. We highlight two important ones below.
\begin{theorem}\label{thm_model2_svd}
The R1-SSVD \eqref{eqssvd2_sdp1} admits an equivalent MISDP formulation:
\begin{align} \label{sdp_two_eq_svd}
w_{\rm SVD}^*&:=\max_{\begin{subarray}{c}
\bm{z}\in Z_{\rm SVD}, \\
\bm X,\bm W_1 , \cdots , \bm W_d \in \S_+^d
\end{subarray}} \Bigg\{ \sum_{i \in [m+n]} \bm{c}_i^{\top} \bm{W}_i \bm{c}_i : \tr(\bm{X}) =1, \bm{X} \succeq \bm{W}_i, \tr(\bm{W}_i) = z_i, \forall i \in [m+n]\Bigg\}
-\sigma_{\max}(\bm A),
\end{align} 
where $\overline{\bm{A}}^\# = \bm{C}^{\top}\bm{C}$ denotes the Cholesky factorization of $\overline{\bm{A}}^\#$ with $\bm C\in \Re^{d\times (m+n)}$, $d$ is the rank of $\overline{\bm{A}}^\#$, and $\bm c_i \in \Re^d$ denotes the $i$-th column vector of matrix $\bm{C}$ for each $i\in [m+n]$.
\end{theorem}

\begin{restatable}{theorem}{themonessvd}\label{thm_model2_relax_svd} 
The continuous relaxation value $\overline{w}_{\rm SVD1}$ of formulation \eqref{sdp_two_eq_svd} satisfies
$$w_{\rm SVD}^* \le \overline{w}_{\rm SVD1} \le \sqrt{mnk_1^{-1}k_2^{-1}} w_{\rm SVD}^*.$$
\end{restatable}
\begin{proof}
See Appendix \ref{proof_thm_model2_relax_svd}. \qed
\end{proof}


\subsection{MISDP Formulation (II)}

Since the results in \Cref{sec:sdpsecond} do not rely on the positive semi-definiteness of matrix $\bm{A}$, they can be directly extended to R1-SSVD \eqref{eqssvd2}. 

We first illustrate a naive MISDP for R1-SSVD \eqref{eqssvd2} based on Part (ii) in \Cref{lem:r1ssvd}.

\begin{proposition} \label{svdthem_eq}
The R1-SSVD \eqref{eqssvd2} is equivalent to the following MISDP formulation:
\begin{align} \label{svd_sdp_one}
 w_{\rm SVD}^* := \frac{1}{2} \max_{\begin{subarray}{c}
\bm{z} \in Z_{\rm SVD},
 \bm{X}\in \S_+^{m+n}
\end{subarray}} \Bigg \{ \tr(\overline{\bm A}\bm{X}) : &\sum_{j\in [m]} X_{jj}=1, \sum_{j\in [m+1, m+n]} X_{jj}=1, X_{ii} \le z_{i}, \forall i \in [m+n]\Bigg \}.
\end{align}
\end{proposition}
The R1-SSVD formulation \eqref{svd_sdp_one} is rather weak and its continuous relaxation value is equal to $\sigma_{\max}(\bm{A})$. 
Fortunately, we can derive two types of valid inequalities from strengthening it as below. 

\begin{restatable}{lemma}{svdlemineq}\label{svdlem_ineq}
For R1-SSVD \eqref{svd_sdp_one}, the following second-order conic inequalities are valid:
\begin{enumerate}[(i)]
	\item $ \sum_{j \in [m]} X_{ij}^2 \le z_i X_{ii}, \sum_{j \in [m+1, m+n]} X_{ij}^2 \le z_i X_{ii}$ for all $i \in [m+n]$; and
	\item $(\sum_{j \in [m]} | X_{ij}|)^2\le k_1 X_{ii} z_i, (\sum_{j \in [m+1,m+n]} | X_{ij}|)^2 \le k_2 X_{ii} z_i$ for all $i \in [m+n]$.
\end{enumerate}
\end{restatable}
\begin{proof} 
See Appendix \ref{proof_svdlem_ineq}. \qed
\end{proof}
The MISDP formulation for R1-SSVD \eqref{svd_sdp_one} can be strengthened by adding these valid inequalities. Similar to \Cref{them_contsdp_one_gap}, we provide the optimality gap of its continuous relaxation value as below.
\begin{restatable}{theorem}{svdthemone} \label{svdthem_contsdp_one}
The continuous relaxation value $\overline{w}_{\rm SVD2}$ of R1-SSVD \eqref{svd_sdp_one} with the inequalities in \Cref{svdlem_ineq} yields an optimality gap at most $\min\{ \sqrt{k_1k_2},mnk_1^{-1}k_2^{-1}\}$, i.e., 
$$w_{\rm SVD}^*\le \overline{w}_{\rm SVD2} \le \min\left\{ \sqrt{k_1k_2}, \sqrt{mnk_1^{-1}k_2^{-1}}\right \} w_{\rm SVD}^*.$$
\end{restatable}

\subsection{An MILP Formulation with Arbitrary Accuracy}

Similarly, we can develop an MILP formulation with arbitrary accuracy based on the Cholesky decomposition of matrix $\overline{\bm{A}}^{\#}$ in R1-SSVD \eqref{eqssvd2_sdp1}. The proofs are similar to \Cref{sec:MILP} and are thus omitted.
\begin{theorem}\label{them_milp_svd}
Given a threshold $\epsilon>0$ and lower and upper bounds of the optimal R1-SSVD, $w_L, w_U$, the following MILP is $O(\epsilon)$-approximate to R1-SSVD \eqref{eqssvd2_sdp1}, i.e., $\epsilon\leq \hat w(\epsilon)- w^* \le \epsilon\sqrt{d}$:
\begin{equation}\label{eq:spca_milp_svd}
\allowdisplaybreaks\begin{aligned} 
\hat{w}(\epsilon) := &\max_{\begin{subarray}{c}
w, \bm{z} \in Z_{\rm SVD}, \bm y, \bm{\alpha},\bm{x}, , \bm \delta,\bm \mu,\bm \sigma
\end{subarray}} w-\sigma_{\max}(\bm{A})\\
\text{\rm s.t.}\quad
& {\bm{x} = \bm \delta_{i1}+\bm \delta_{i2}, ||\bm \delta_{i1}||_{\infty}\le z_i, ||\bm \delta_{i2}||_{\infty}\le 1- z_i, \forall i \in [m+n]}, \\
&{ \bm x = \sum_{j \in [d]} \bm \sigma_{j}, ||\bm \sigma_{j}||_{\infty}\le y_{j}, \sigma_{jj} = y_{j}, \forall j \in [d], \sum_{j \in [d]} y_{j}=1}, \\
&{\bm x= \bm \mu_{\ell 1} +\bm \mu_{\ell 2}, ||\bm \mu_{\ell 1}||_{\infty}\le \alpha_{\ell}, ||\bm \mu_{\ell 2}||_{\infty}\le 1- \alpha_\ell, \forall \ell\in [L]}, \\
&w=w_U - (w_U-w_L)\bigg(\sum_{i\in [L]} 2^{-i}\alpha_i\bigg), \\
 &{\bigg|\bigg|\sum_{i \in [m+n]}\bm{c}_i \bm{c}_i^{\top}\bm\delta_{i1} - w_U \bm{x} + (w_U-w_L)\sum_{i\in [L]} 2^{-i}\bm \mu_{i1}\bigg|\bigg|_{\infty} \le \epsilon }, \\
& \bm \alpha \in \{0,1\}^L, \bm y \in \{0,1\}^d,
\end{aligned}
\end{equation}
where $L:= \lceil \log_2 (\epsilon/(w_U-w_L))\rceil $.
\end{theorem}

\begin{theorem}\label{thm_approx_milp_svd}
Given a threshold $\epsilon>0$, let $\overline{w}_{\rm SVD3}(\epsilon)$ denote the optimal value of MILP formulation \eqref{eq:spca_milp_svd} by relaxing the binary variables $\bm{z}$ to be continuous. Then we have 
\begin{align*}
 \overline{w}_{\rm SVD3}(\epsilon)\leq & \sqrt{\frac{mn}{k_1k_2}}\bigg[\min\bigg\{(k_1+k_2)\frac{\sqrt{d}+1}{2},\\
 &\frac{m+n}{k_1+k_2} \sqrt{d} + (m+n-k_1-k_2)\frac{\sqrt{d}+1}{2}\bigg\}-1\bigg]w^*_{\rm SVD} +\epsilon\sqrt{d}. 
\end{align*}

\end{theorem}

\subsection{Approximation Algorithms for R1-SSVD}
We will investigate three approximation algorithms for R1-SSVD \eqref{ssvd}: truncation algorithm, greedy algorithm, and local search algorithm.

\subsubsection{Truncation algorithm} 
The approximation algorithm in \cite{chan2016approximability} via truncation is known so far with the best approximation ratio $O(n^{-1/3})$ for SPCA. We show that a similar truncation also works for R1-SSVD. 

First, we define the truncation operator as below.
\begin{definition}[Normalized Truncation]\label{def_trunc}
Given a vector $\bm x \in \Re^n$ and an integer $s\in [n]$, vector $\hat{\bm x}$ is an $s$-truncation of $\bm x$ if
\begin{align*}
&\hat{x}_{i}=
\begin{cases}
|{x}_{i}|, \ \ \ \ &\text{\rm if $|x_i|$ is one of the $s$ largest absolute entries of $\bm{x}$} \\
0, \ \ \ \ &\text{\rm otherwise}
\end{cases}
\end{align*}
for each $i\in [n]$. The normalized $s$-truncation of $\bm x$ is defined as $\hat{\bm x}:=\hat{\bm x}/\|\hat{\bm x}\|_2$, which is normalized to be of unit length.
\end{definition}

Then the truncation algorithm for R1-SSVD has the following two steps:\par
\noindent\textbf{(i) Truncation in the standard basis:} For each $i\in [n]$, let $\hat{\bm{u}}_i \in \Re^m$ be the normalized $k_1$-truncation on the $i$-th column vector of $\bm A$, and for each $j\in [m]$, let $\hat{\bm{v}}_j \in \Re^n$ be the normalized $k_2$-truncation on the $j$-th row vector of $\bm A$. Clearly, $\hat{\bm{u}}_i$ and $\hat{\bm{v}}_j$ are feasible to R1-SSVD \eqref{ssvd};



\noindent\textbf{(ii) Truncation in the eigen-space basis:}
Let $\bm{v}_1$ and $\bm{u}_1$ denote the right and left eigenvectors of $\bm{A}$ corresponding to the largest singular value.
We then define the vector $\hat{\bm{u}}_1$ as the normalized $k_1$-truncation on $\bm{u}_1$ and define $\hat{\bm{v}}_1$ as the normalized $k_2$-truncation of the vector $\bm{A}^\top \hat{\bm u}_1$. It is clear that $({\hat{\bm{u}}_1}, {\hat{\bm{v}}_1} )$ is also feasible to R1-SSVD \eqref{ssvd}.


The approximation results of the truncation procedure are summarized below.

\begin{restatable}{theorem}{thmtruncation}\label{thm_truncation}
	For R1-SSVD \eqref{ssvd}, the truncation algorithm yields an approximation ratio 
	$$\max \left\{ \sqrt{k_1^{-1}} ,\sqrt{k_2^{-1}},\sqrt{k_1k_2m^{-1}n^{-1}} \right\}.$$
	In particular, the approximation ratio is $O(n^{-1/3})$ when $k_1\approx k_2$ and $m\approx n$.
\end{restatable}
\begin{proof}
See Appendix \ref{proof_thm_truncation}. \qed
\end{proof}

\subsubsection{Greedy and Local Search Algorithms}
We design the greedy and local search algorithms according to the following equivalent combinatorial formulation of R1-SSVD \eqref{ssvd}
\begin{align} \label{eqssvdcom}
w_{\rm SVD}^* :=\max_{S_1\subseteq [m], S_2\subseteq [n]} &\bigg\{ \sigma_{\max}({\bm{A}}_{S_1,S_2}): |S_1|=k_1,|S_2|=k_2 \bigg\}.
\end{align}
Different from SPCA \eqref{spcacom}, the R1-SSVD \eqref{eqssvdcom} maximizes the largest singular value of any $k_1 \times k_2$ submatrix rather than that of any size $k$-principal submatrix. Therefore, to solve R1-SSVD \eqref{eqssvdcom}, we adapt the greedy \Cref{alg:greedy} or the local search \Cref{alg:localsearch} considering selecting a row and/or a column at each iteration.
 
Specifically, for the greedy algorithm, let two subsets $S_1, S_2$ denote the index sets of the selected columns and rows, respectively. We first initialize the greedy algorithm by selecting the entry of $\bm A$ that takes the largest absolute value.
Then, we add one element into each subset at each iteration, which maximizes the largest singular value of the obtained submatrix, unless we are not able to. Next, we continue to selection one row (or one column) at each iteration, until we reach a $k_1 \times k_2$ submatrix. The detailed implementation can be found in \Cref{alg:svd_greedy}. 

Given an initial feasible solution $(S_1, S_2)$ to R1-SSVD \eqref{eqssvdcom}, the adapted local search algorithm performs the swapping procedure on both $S_1$ and $S_2$ (see \Cref{alg:svd_localsearch} for details) simultaneously.

\begin{algorithm}[htbp]
	\caption{Greedy Algorithm for R1-SSVD \eqref{eqssvdcom}}
	\label{alg:svd_greedy}
	\begin{algorithmic}[1]
		\State \textbf{Input:} $m \times n$ matrix $\bm{A} \succeq 0$, integers $k_1 \in [m]$, $k_2\in [n]$
		\State Let $\hat{S}_1 := \emptyset$ and $\hat{S}_2 := \emptyset$ denote the selected rows and columns, separately
		
		\State Compute $j_1^*, j_2^* \in \argmax_{j_1 \in[m],j_2 \in[n] }\left\{|(\bm A_{\{j_1\}, \{j_2\}} |\right\}$
		\State Add $j_1^*, j_2^*$ to sets $\hat{S}_1$ and $\hat{S}_2$, separately
		
		\For{$\ell = 2, \cdots, \max\{k_1,k_2\}$}
		\If{$\ell \le \min\{k_1,k_2\}$}
		\State Compute
		$j_1^* \in \argmax_{j_1 \in[m]\setminus \hat{S}_1 }\left\{\sigma_{\max}\left(\bm A_{\hat S_1\cup\{j_1\}, \hat S_2 }\right)\right\}$ and add $j_1^*$ to set $\hat{S}_1$ 
		
		\State Compute $j_2^* \in \argmax_{j_2 \in[n]\setminus \hat{S}_2 }\left\{\sigma_{\max}\left(\bm A_{\hat S_1, \hat S_2\cup\{j_2\} }\right)\right\}$ and add $j_2^*$ to set $\hat{S}_2$
		\ElsIf {$k_1\leq k_2$}
		\State Compute
		$j^*_2 \in \argmax_{j_2 \in[n]\setminus \hat{S}_2}\left\{\sigma_{\max}\left(\bm A_{S_1, S_2\cup\{j_2\}} \right)\right\}$ and add $j^*_2$ to set $\hat{S}_2$
		\Else
		\State Compute
		$j^*_1 \in \argmax_{j_1 \in[m]\setminus \hat{S}_1}\left\{\sigma_{\max}\left(\bm A_{\hat S_1\cup\{j_1\}, \hat S_2} \right)\right\}$ and add $j^*_1$ to set $\hat{S}_1$
		\EndIf
		\EndFor
		
		\State \textbf{Output:} $\hat S_1, \hat S_2$
	\end{algorithmic}
\end{algorithm}

The following results illustrate the theoretical performance guarantees of the two algorithms for R1-SSVD and show that the approximation ratios are both tight. 
\begin{restatable}{theorem}{thmsvdalg}\label{thm_svdgreedy}
For the greedy \Cref{alg:svd_greedy} and the local search \Cref{alg:svd_localsearch}, we have (i) both algorithms achieve a $(\sqrt{k_1k_2})^{-1}$-approximation ratio of R1-SSVD \eqref{eqssvdcom}, and (ii) the ratio is tight.
\end{restatable}
\begin{proof}
See Appendix \ref{proof_thm_svdgreedy}. \qed
\end{proof}

\begin{algorithm}[htbp]
	\caption{Local Search Algorithm for R1-SSVD \eqref{eqssvdcom}}
	\label{alg:svd_localsearch}
	\begin{algorithmic}[1]
		\State \textbf{Input:} $m\times n$ matrix $\bm{A} \succeq 0$ and integers $k_1 \in [m]$, $k_2 \in [n]$
		\State Initialize a size-$k_1$ subset $\hat{S}_1 \subseteq [m]$ and a size-$k_2$ subset $\hat{S}_2 \subseteq [n]$ 
		\Do
		\For{each pair {$(i_1,j_1, i_2, j_2) \in \hat{S}_1 \times ([m]\setminus \hat{S}_1)\times \hat S_2 \times([n]\setminus \hat S_2) $}}
		\If{$\sigma_{\max} \left(\bm A_{S_1\cup \{j_1\}\setminus \{i_1\}, S_2\cup \{j_2\} \setminus \{i_2\}} \right) > \sigma_{\max} \left(\bm A_{S_1, S_2}\right)$}
		\State Update $\hat{S}_1 := \hat{S}_1 \cup \{j_1\} \setminus \{i_1\}$, $\hat{S}_2 := \hat{S}_2 \cup \{j_2\} \setminus \{i_2\}$
		\EndIf
		\EndFor
		\doWhile{there is still an improvement}
		\State \textbf{Output:} $\hat S_1$, $\hat S_2$
	\end{algorithmic}
\end{algorithm}

\section{Extension to Sparse Fair PCA}\label{sec:sfpca}
In this section, we study the Sparse Fair PCA (SFPCA) and show its approximate MISDP formulation. The fair PCA has been recently studied in the literature (see, e.g., \cite{samadi2018price,tantipongpipat2019multi}). The goal of SFPCA is to seek the best principal submatrices of multi-group covariance matrices to achieve the relatively similar objective values among different groups. 


Suppose there are $s$ groups and their corresponding covariance matrices are $\{\bm{A}_i\}_{i\in [s]}$. Then the SFPCA can be formulated as
\begin{align}
\label{sfpca}
w_{F}^* := \max_{\bm{x}} \left \{\min_{i\in S}\bm{x}^{\top}\bm{A}_i\bm{x}: ||\bm{x}||_2 = 1, ||\bm{x}||_0\le k \right \}.
\end{align}
By introducing binary variables $\bm{z}$ and linearizing the objective function, we obtain
\begin{align}
\label{sfpca_binary}
w_{F}^* := \max_{w,\bm{x}, \bm{z}\in Z} \left \{w : w\le \bm{x}^{\top}\bm{A}^i\bm{x}, \forall i \in [s], ||\bm{x}||_{2}=1,-z_{i} \le x_{i} \le z_{i}, \forall i \in [n] \right \}.
\end{align}

As the SFPCA \eqref{sfpca_binary} is quite different from SPCA, it is not surprising that the results in \Cref{sec:sdpone} and \Cref{sec:MILP} do not apply to SFPCA \eqref{sfpca_binary}. Fortunately, the results in \Cref{sec:sdpsecond} do provide an interesting upper bound for SFPCA \eqref{sfpca_binary}, which can be exact when there are $s=2$ groups of covariance matrices. Introducing a rank-one positive semi-definite matrix variable $\bm{X}\in \S_+^n$ such that $\bm X\succeq \bm{x}\bm{x}^\top$, dropping the rank-one restriction, and adding the valid inequalities in \Cref{them_eq_pca_stronger}, the problem \eqref{sfpca_binary} can be upper bounded by
\begin{align}
\label{sfpca_sdp}
 \overline{w}_{F} := \max_{w,\bm{X}, \bm{z}\in Z} &\bigg\{w : w\le \tr (\bm{A}^i \bm{X}), \forall i \in [s], \tr (\bm{X})=1,\notag\\
 &\sum_{j\in [n]} X_{ij}^2 \le X_{ii} z_i, \bigg(\sum_{j \in [n]} | X_{ij}| \bigg)^2 \le k X_{ii}z_i,\forall i \in [n]\bigg\}.
\end{align}

The following result shows that if $s=2$, then the approximation \eqref{sfpca_sdp} is exact, otherwise, it provides an upper bound of SFPCA \eqref{sfpca_binary}.
\begin{proposition}\label{prop_sfpca} For the MISDP formulation \eqref{sfpca_sdp}, we have
\begin{enumerate}[(i)]
\item The optimal value of MISDP formulation \eqref{sfpca_sdp} provides an upper bound of SFPCA \eqref{sfpca_binary}, i.e., $\overline{w}_{F}\geq w_F^*$. Also, when $s=2$, the formulation \eqref{sfpca_sdp} becomes exact, i.e., $\overline{w}_{F}=w_F^*$; and
\item There exists an optimal solution $(w^*,\bm{X}^*, \bm{z}^*)$ of of MISDP \eqref{sfpca_sdp} such that the rank of $\bm{X}^*$ is at most $1+\lfloor \sqrt{2s+9/4}-3/2\rfloor$.
\end{enumerate}
\end{proposition}
\begin{proof}
\begin{enumerate}[(i)]
\item
It is clear that $\overline{w}_{F}\geq w_F^*$ since we drop the rank-one restriction on $\bm{X}$ of MISDP formulation \eqref{sfpca_sdp}. On the other hand, for the case of $s= 2$, theorem 1.1 in \cite{tantipongpipat2019multi} shows that for any feasible solution $(w,\bm{X},\bm{z})$, there exists a rank-one semi-definite matrix $\hat{\bm X}$ such that the new solution $(w,\hat{\bm{X}},\bm{z})$ is also feasible and achieves the same objective value. Thus, we must have $\overline{w}_{F}=w_F^*$;

\item Suppose $(w,\bm{X}, \bm{z})$ denotes an optimal solution of MISDP \eqref{sfpca_sdp}. Let $S=\{i\in [n]: z_i=1\}$. Then according to theorem 1.7 in \cite{tantipongpipat2019multi}, there exists a  semi-definite matrix $\hat{\bm X}$ of the rank at most $1+\lfloor \sqrt{2s+9/4}-3/2\rfloor$ such that the new solution $(w,\hat{\bm{X}},\bm{z})$ is also optimal. 
\end{enumerate}

\qed
\end{proof}

\Cref{prop_sfpca} shows that two-group SFPCA \eqref{sfpca_binary} admits an MISDP representation, while MISDP formulation \eqref{sfpca_sdp} provides a low-rank solution in general for SFPCA when $s>2$. It is worthy of mentioning that the results in \Cref{prop_sfpca} work for any convex fairness measure.

\section{Conclusion} \label{sec:conclusion}
In practice, to tune the parameter $k$ via cross-validation, our developed greedy and local search algorithms can be quickly warm started from solution procedure in the previous iterations. We anticipate that the theoretical optimality gaps of three exact formulations for SPCA and R1-SSVD are not tight and can be further strengthened. The analysis of the optimality gap of sparse fair PCA requires new techniques, which can be an exciting research direction. Also, it might be desirable to study robust sparse PCA when the datasets are noisy or contain outliers.

\bibliography{reference.bib}
\newpage
\titleformat{\section}{\large\bfseries}{\appendixname~\thesection .}{0.5em}{}
\begin{appendices}
	 \section{Proofs}\label{proofs}
\subsection{Proof of \Cref{lemspca}} \label{proof:lemspca}
\lemmaspca*
	\begin{proof}
	\noindent\textbf{Part (i)}
	Given a size-$k$ set $S\subseteq [n]$, the maximization problem $$\max_{\bm{x} \in \Re^n} \left \{\bm{x}^{\top}\bm{A}\bm{x}: {||\bm{x}||_{2}=1}, x_i=0, \forall i \notin S \right \}$$ reduces to
	$$\max_{\bm{x} \in \Re^k} \left \{\bm{x}^{\top}\bm{A}_{S,S}\bm{x}: {||\bm{x}||_{2}=1}\right \},$$
	which is exactly the definition of the largest eigenvalue of principal submatrix $\bm{A}_{S,S}$.
	
	\noindent\textbf{Part (ii)} According to Part (i), it is sufficient to show that $v^* = \hat v$, where $v^*, \hat v$ are defined as
	\begin{align}
	v^*:&=\max_{\bm{X} \in \S_+^k} \left \{ \tr(\bm{A}_{S,S}\bm{X}) : \tr(\bm{X})=1 \right \}, \label{eq_v_star}\\ 
	\hat{v}:&=\max_{\bm{x} \in \Re^k} \left \{\bm{x}^{\top}\bm{A}_{S,S}\bm{x}: {||\bm{x}||_{2}=1}\right \}. \label{eq_v_hat} 
	\end{align}
	
	First, we must have $v^* \ge \hat v$. Indeed, for any feasible $\bm x\in \Re^k$ to problem \eqref{eq_v_hat} such that $||\bm{x}||_{2}=1$, we can construct a positive semi-finite matrix by $\bm X=\bm x\bm x^{\top}$, which is feasible to problem \eqref{eq_v_star} and yields the same objective value. 
	
	Second, to prove $\hat v \ge v^*$, we let $\bm X^* \in \S_+^k$ denote an optimal solution to problem \eqref{eq_v_star} and $\bm X^* =\sum_{i\in [k]}\lambda_i \bm q_i \bm q_i^{\top}$ denote its spectral decomposition. Since $\tr(\bm{X}^*)=1$ and $\bm X^* \in \S_+^k$, the eigenvalues must satisfy $\sum_{i\in [k]} \lambda_i=1$ and $\lambda_i\geq 0$ for each $i\in [k]$. Thus, the optimal value $v^*$ of problem \eqref{eq_v_star} is equal to
	$$v^* = \tr(\bm{A}_{S,S}\bm{X}^*) = \sum_{i\in [k]} \lambda_i \bm q_i^{\top} \bm A_{S,S}\bm q_i \le \max_{i\in [k]} \bm q_i^{\top} \bm A_{S,S}\bm q_i \le \hat v,$$
	where the inequality is due to $\sum_{i\in [k]} \lambda_i=1$ and $\lambda_i\geq 0$ for each $i\in [k]$.
	
	\noindent\textbf{Part (iii)} For a positive semi-definite matrix $\bm A$, let $\bm{A} = \bm{C}^{\top}\bm{C}$ denote the Cholesky factorization of $\bm{A}$ and $\bm C \in \Re^{d\times n}$, thus we have
	\[\lambda_{\max}(\bm{A}_{S,S}) = \lambda_{\max}(\bm{C}^{\top}_{S} \bm{C}_{S}) = \lambda_{\max}(\bm{C}_{S} \bm{C}_{S}^{\top} ),\]
	where the second equality is because for any matrix, its largest singular value is equal to that of its transpose.
	\qed
\end{proof}
\subsection{Proof of \Cref{thm_H1}} \label{proof_thm_H1}
\thmHone*
	\begin{proof} 
\textbf{Part (i).} We split the proof of strong duality into two cases depending on whether $\bm z$ is a relative interior point of set $\overline{Z}$ or not.
	\begin{enumerate}[{Case }a.]
		\item We will first prove the result by assuming that $\bm{z}$ is in the relative interior of set $\overline{Z}$, i.e., $0<z_i<1$ for each $i\in [n]$.
		For the inner maximization problem in \eqref{sdp_two_eq_h1}, we dualize the constraint $\bm{X} \succeq \bm{W}_i, \tr(\bm{W}_i) = z_i$ with Lagrangian multiplier $\bm{Q}_i\in \S_+^d$ and $\mu_i$ for each $i\in [n]$. Note that the constraints $\bm{X} \succeq \bm{W}_i,\tr(\bm{W}_i) = z_i$ for each $i\in [n]$ and $\bm X, \bm W_1 , \cdots, \bm W_n \in \S_+^d$ can be always strictly satisfied since $0<z_i<1$. Thus, according to the strong duality of general conic program (see, e.g., Theorem 1.4.4 in \cite{ben2001lectures}), function $H_1(\bm{z})$ can be rewrite as
		\begin{align}\label{eq_discrbi_H12}
\min_{ \bm{\mu},\bm Q_1,\cdots,\bm Q_n\in \S_+^d} &\max_{
			\bm X,\bm W_1 , \cdots , \bm W_n \in \S_+^d} \Bigg\{ \sum_{i \in [n]} \bm{c}_i^{\top} \bm{W}_i \bm{c}_i + \sum_{i \in [n]} \tr\left(\bm{Q}_i(\bm{X}-\bm{W}_i) \right) + \sum_{i \in [n]} \mu_i \left(z_i-\tr(\bm{W}_i) \right) :\tr(\bm{X}) =1\Bigg\}.
		\end{align}
		Then the inner maximization problem \eqref{eq_discrbi_H12} over $\bm W_i$ for each $i \in [n]$ and $\bm X$ yields
		\begin{align*}
		&\max_{\bm{W}_i \in \S^d_+} \tr \left((\bm{c}_i\bm{c}_i^{\top}-\bm{Q}_i -\mu_{i} \bm{I}_d )\bm{W}_i\right) = \begin{cases}
		0,&\bm{c}_i\bm{c}_i^{\top} \preceq \bm{Q}_i + \mu_{i} \bm{I}_d , \\
		\infty, &\textrm{ otherwise}.
		\end{cases} \\
		&\max_{\bm X \in \S^d_+} \bigg\{ \tr\bigg(\bigg(\sum_{i\in [n]} \bm Q_i\bigg)\bm X\bigg): \tr (\bm X)=1 \bigg\} = \lambda_{\max}\bigg(\sum_{i\in [n]} \bm Q_i\bigg),
		\end{align*}
		where the second identity is due to Part(ii) of \Cref{lemspca}. 
		
		Thus, problem \eqref{eq_discrbi_H12} can be simplified as
		\begin{align}\label{eq_discrbi_H13}
		H_1(\bm{z})=\min_{ \bm{\mu},\bm Q_1,\cdots,\bm Q_n\in \S_+^d} &\bigg\{\lambda_{\max}\bigg(\sum_{i\in [n]} \bm Q_i \bigg) + \sum_{i \in [n]} \mu_i z_i: \bm{c}_i\bm{c}_i^{\top} \preceq \bm{Q}_i + \mu_{i} \bm{I}_d , \forall i \in [n] \bigg\}.
		\end{align}
		
We show that for the minimization problem \eqref{eq_discrbi_H13}, any optimal solution $(\bm{\mu},\bm Q_1,\cdots,\bm Q_n)$ must satisfy $0\le \mu_{i}\le \|\bm{c}_i\|_2^2$ for each $i \in [n]$. We prove it by contradiction. Suppose that there exits an optimal solution $(\bm{\mu},\bm Q_1,\cdots,\bm Q_n)$ to the problem \eqref{eq_discrbi_H13} such that $\mu_j<0$ for some $j \in [n]$. Then, we can construct a new feasible solution $(\overline{\bm{\mu}},\overline{\bm Q}_1,\cdots,\overline{\bm Q}_n)$, which is exactly equal to $(\bm{\mu},\bm Q_1,\cdots,\bm Q_n)$ except
	\[\overline{\mu}_j = 0, \overline{\bm Q}_j = \bm Q_j+\mu_j \bm{I}_d.\]
The new solution yields the objective value 
\[H_1(\bm{z})+\mu_j - \mu_j z_j = H_1(\bm{z})+\mu_j(1-z_j) < H_1(\bm{z}), \]
which is a contradiction to the optimality of $(\bm{\mu},\bm Q_1,\cdots,\bm Q_n)$.
Similarly, suppose that there exits an optimal solution $(\bm{\mu},\bm Q_1,\cdots,\bm Q_n)$ to the problem \eqref{eq_discrbi_H13} such that $\mu_j>\|\bm{c}_i\|_2^2$ for some $j \in [n]$. Similarly, we can arrive at a contradiction by defining a new feasible solution $(\overline{\bm{\mu}},\overline{\bm Q}_1,\cdots,\overline{\bm Q}_n)$, which is exactly equal to $(\bm{\mu},\bm Q_1,\cdots,\bm Q_n)$ except $\overline{\mu}_j = \|\bm{c}_i\|_2^2$.

Therefore, \eqref{eq_discrbi_H13} can be reduced to \eqref{eq_discrbi_H14}.

		\item Now we consider the case that $\bm{z}$ is not in the relative interior of $\overline{Z}$ and define two sets $T_0:=\{i\in [n]:z_i=0\}$ and $T_1:=\{i\in [n]:z_i=1\}$. Thus, at least one of the two sets is not empty. In this case, we first observe that $H_1(\bm{z})$ in \eqref{sdp_two_eq_h1} is equivalent to
		\begin{align} \label{sdp_two_eq_noint}
		H_1(\bm{z}):=\max_{
			\bm X,\bm W_1 , \cdots , \bm W_d \in \S_+^d} & \Bigg\{ \sum_{i \in [n]\setminus (T_0\cup T_1)} \bm{c}_i^{\top} \bm{W}_i \bm{c}_i+ \sum_{i \in T_1} \bm{c}_i^{\top} \bm{X} \bm{c}_i: \tr(\bm{X}) =1,\notag\\
		&\bm{X} \succeq \bm{W}_i, \tr(\bm{W}_i) = z_i, \forall i \in [n]\setminus (T_0\cup T_1)\Bigg\}.
		\end{align}
		Next, applying the same procedure as Case a., we have
		\begin{align}\label{eq_discrbi_H14_noint}
		H_1(\bm{z})=\min_{ \bm{\mu},\{\bm Q_i\}_{i\in [n]\setminus (T_0\cup T_1)}\subseteq  \S_+^d} &\bigg\{\lambda_{\max}\bigg(\sum_{i\in [n]\setminus (T_0\cup T_1)} \bm Q_i +\sum_{i\in T_1} \bm{c}_i\bm{c}_i^{\top}\bigg) + \sum_{i \in [n]\setminus (T_0\cup T_1)} \mu_i z_i: \notag\\
		&\bm{c}_i\bm{c}_i^{\top} \preceq \bm{Q}_i + \mu_{i} \bm{I}_d , 0\le \mu_{i}\le \|\bm{c}_i\|_2^2 , \forall i \in [n]\setminus (T_0\cup T_1)\bigg\}.
		\end{align}
To show the equivalence between \eqref{eq_discrbi_H14_noint} and \eqref{eq_discrbi_H14}, it remains to prove that
		\begin{align}\label{eq_discrbi_H14_noint_new}
		\hat{H}_1(\bm{z})=\min_{ \bm{\mu},\{\bm Q_i\}_{i\in [n]}\subseteq \S_+^d} &\bigg\{\lambda_{\max}\bigg(\sum_{i\in [n]} \bm Q_i \bigg) + \sum_{i \in [n]} \mu_i z_i: \notag\\
		&\bm{c}_i\bm{c}_i^{\top} \preceq \bm{Q}_i + \mu_{i} \bm{I}_d , 0\le \mu_{i}\le \|\bm{c}_i\|_2^2 , \forall i \in [n]\bigg\}.
		\end{align}

First, given any feasible solution $(\bm{\mu},\{\bm Q_i\}_{i\in [n]\setminus (T_0\cup T_1)})$ to the problem \eqref{eq_discrbi_H14_noint}, let us augment it by setting $\bm{Q}_i=\bm0,\mu_i=\|\bm{c}_i\|_2^2$ for each $i\in T_0$ and $\bm{Q}_i=\bm{c}_i\bm{c}_i^{\top},\mu_i=0$ for each $i \in T_1$. Then $(\bm{\mu},\{\bm Q_i\}_{i\in [n]})$ is feasible to the problem \eqref{eq_discrbi_H14_noint_new} with the same objective value. Thus, we have $\hat{H}_1(\bm{z})\leq H_1(\bm{z})$.

On the other hand, given any feasible solution $(\bm{\mu},\{\bm Q_i\}_{i\in [n]})$ to the problem \eqref{eq_discrbi_H14_noint_new}, then $(\bm{\mu},\{\bm Q_i\}_{i\in [n]\setminus (T_0\cup T_1)})$ is feasible to the problem \eqref{eq_discrbi_H14_noint} a smaller objective value since $\bm{c}_i\bm{c}_i^{\top} \preceq \bm{Q}_i + \mu_{i}$ for each $i\in T_1$. Thus, we have $\hat{H}_1(\bm{z})\geq H_1(\bm{z})$. This completes the proof.
	\end{enumerate}
\textbf{Part (ii).} For any $z \in Z$, let set $S$ denote its support. We then construct a pair of the primal and dual solutions to the maximization problem in \eqref{sdp_two_eq_h1} and its dual \eqref{eq_discrbi_H14} as
	\begin{align*}
&\bm X^* = \bm q_1 \bm q_1^{\top}, \bm W^*_i = \bm X^*, \forall i \in S, \bm W^*_i =0, \forall i \in [n]\setminus S,\\
&\bm Q_i^* = \bm c_i \bm c_i^{\top}, \mu_i=0, \forall i \in S, \bm Q_i^* =0, \mu_i = ||\bm c_i||_2^2, \forall i \in [n] \setminus S,
	\end{align*}
where $\bm q_1$ denote the eigenvector for the largest eigenvalue of matrix $\sum_{i\in S}\bm c_i \bm c_i^{\top}$.

According to the results in \Cref{lemspca}, the above solutions return the same objective value for primal and dual problems, which is $\lambda_{\max}(\sum_{i\in S}\bm c_i \bm c_i^{\top})$. This proves the optimality of the proposed dual solution.
\qed
\end{proof}
\subsection{Proof of \Cref{them_eq_pca}} \label{proof_prop2}
\sdptwoprop*
	\begin{proof}
	\begin{enumerate}[(i)]
		\item To show the equivalence of problem \eqref{sdp_one} and SPCA \eqref{spca2}, we only need to show that for any feasible $\bm{z}\in Z$ with its support $S=\{i: z_i =1\}$, we must have
		\begin{equation}\label{eq_AX_lambda_S}
		\max_{\bm{X}\in \S^n_+} \bigg \{ \tr(\bm{A}\bm{X}) : \tr(\bm{X})=1, X_{ii} \le z_{i}, \forall i \in [n] \bigg \}=\lambda_{\max}(\bm{A}_{SS}).
		\end{equation}
		Indeed, since $\bm{X}$ is a positive semi-definite matrix, thus $ X_{ii}=0$ for each $i\in [n]\setminus S$ implies
		$$X_{ij}=0, \forall (i,j)\notin S\times S.$$
		The left-hand side of \eqref{eq_AX_lambda_S} is equivalent to
		\begin{align*}
		\max_{\bm{X}\in \S^n_+} \bigg \{ \tr(\bm{A}\bm{X}) : \tr(\bm{X})=1, X_{ii} \le z_{i}, \forall i \in [n] \bigg \}= \max_{\bm{X} \in \S_+^k} \left \{ \tr(\bm{A}_{S,S}\bm{X}) : \tr(\bm{X})=1 \right \}=\lambda_{\max}(\bm{A}_{SS}),
		\end{align*}
		where the second equality is due to Part (ii) in \Cref{lemspca}. 
		
		\item The continuous relaxation value of problem \eqref{sdp_one} is
		\begin{align*}
		\overline{w}_3 = \max_{\bm{z} \in \overline{Z}, \bm{X}\in \S^n_+} \bigg \{ \tr(\bm{A}\bm{X}) : \tr(\bm{X})=1, X_{ii} \le z_{i}, \forall i \in [n] \bigg \}.
		\end{align*}
		Since $\tr(\bm{X})=1$, thus the linking constraint $ X_{ii} \le z_{i}$ is redundant for each $i\in [n]$. Hence,
		\begin{align*}
		\overline{w}_3 = \max_{ \bm{X}\in \S^n_+} \bigg \{ \tr(\bm{A}\bm{X}) : \tr(\bm{X})=1 \bigg \}=\lambda_{\max}(\bm{A}),
		\end{align*}
		where the equality is due to Part (ii) in \Cref{lemspca}. \qed
	\end{enumerate}
\end{proof}
\subsection{Proof of \Cref{lem_ineq}} \label{proof_lem_ineq}
\lemineq*
	\begin{proof}
	From the proof of \Cref{them_eq_pca}, there must exists an optimal solution $(\bm{z}^*,\bm X^*)$ of SPCA \eqref{sdp_one} such that $\bm{X}^*$ must be rank-one. Thus, without loss of generality, for any feasible solution $(\bm{z},\bm X)$ of SPCA \eqref{sdp_one}, we can assume that $\bm X = \bm x \bm x^{\top}$, where $(\bm{x},\bm{z})$ is also feasible to SPCA \eqref{spca2}.
	
	Next, we split the proof into two parts.
	\begin{enumerate}[(i)]
		\item Since $\bm X = \bm x \bm x^{\top}$, thus
		$$ \sum_{j \in [n]} X_{ij}^2=\sum_{j\in [n]} x_{i}^2 x_{j}^2 = x_{i}^2\le z_i X_{ii}, \forall i \in [n],$$
		where the last inequality follows from the facts that $ X_{ii} = x_{i}^2 \le z_i$ and $z_i$ is binary for each $i \in [n]$.
		
		\item It is known (see, e.g., \cite{dey2018convex}) that $||\bm x||_1 \le \sqrt{k}$. Thus,
		$$\sum_{j \in [n]} | X_{ij}|= \sum_{j \in [n]}|x_i| |x_j| \le \sqrt{k} |x_i| \le \sqrt{k} \sqrt{ X_{ii} z_i},$$
		where the second inequality is due to the facts that $ X_{ii} = x_{i}^2 \le z_i$ and $z_i$ is binary for each $i \in [n]$.\qed
	\end{enumerate}
\end{proof}
\subsection{Proof of \Cref{them_milp}} \label{proof_them_milp}
\themmilp*
\begin{proof} Throughout the proof, we use indices $i \in [n]$, $j\in [d]$, and $\ell \in [m]$ to denote the elements of three different dimensional vectors, respectively.
	To construct the MILP by SPCA \eqref{spcacom3_eigen} and show the approximation accuracy, we split the proof into four steps.
	\begin{enumerate}[{\bf Step} 1.]
		\item Linearize the bilinear terms $\{z_i\bm{x}\}_{i\in [n]}$ in \eqref{spcacom3_eigen}. This can be done by introducing two copies $\bm \delta_{i1},\bm \delta_{i2}$ of vector $\bm{x}$ for each $i\in [n]$ such that
		\[{\bm{x} = \bm \delta_{i1}+\bm \delta_{i2}, ||\bm \delta_{i1}||_{\infty}\le z_i, ||\bm \delta_{i2}||_{\infty}\le 1- z_i, \forall i \in [n]}, \sum_{i \in [n]}z_i\bm{c}_i\bm{c}_i^\top \bm{x}=\sum_{i \in [n]}\bm{c}_i\bm{c}_i^\top \bm \delta_{i1}.\]
		
		\item Linearize the nonconvex constraint $\|\bm{x}\|_{\infty}=1$. We first observe that due to symmetry, $\|\bm{x}\|_{\infty}=1$ can be equivalently written as a disjunction with $d$ sets as below
		\[\cup_{j\in [d]}\big\{\bm{x} \in \Re^d: x_j=1,\|\bm{x}\|_{\infty}\leq 1 \big\}.\]
		Next, for each $j\in d$, we introduce a binary variable $y_j=1$ indicating the $j$-th set is active and 0, otherwise, and then create a copy $\bm \sigma_{j} \in \Re^d$ of variable $\bm{x}$ such that
		\[{ \bm x = \sum_{j \in [d]} \bm \sigma_{j}, ||\bm \sigma_{j}||_{\infty}\le y_{j}, \sigma_{jj} = y_{j}, \forall j \in [d], \sum_{j \in [d]} y_{j}=1},\bm{y}\in \{0,1\}^d.\]
		
		\item Approximate and linearize bilinear term $w\bm{x}$. We first approximate variable $w$ using $m$ binary variables $\bm{\alpha}\in \Re^m$ with $m:= \lceil \log_2 ((w_U-w_L)/\epsilon)\rceil $. Thus, we have
		\[w\approx w_U - (w_U-w_L)\bigg(\sum_{\ell\in [m]} 2^{-\ell}\alpha_{\ell}\bigg)\]
		with approximation accuracy at most $(w_U-w_L)/2^{m}\leq \epsilon$. The bilinear term $w\bm{x}$ is now approximated by
		\begin{align}\label{eq_approx}
	w\bm{x}\approx w_U\bm{x} - (w_U-w_L)\bigg(\sum_{\ell\in [m]} 2^{-\ell}\alpha_{\ell}\bm{x}\bigg).
		\end{align}
		With binary variables $\bm \alpha$, the resulting bilinear terms $\{\alpha_{\ell}\bm{x}\}_{\ell\in [m]}$   can be further linearized following the same arguments as Step 2, i.e., 
		\begin{align*}
		&{\bm x= \bm \mu_{\ell 1} +\bm \mu_{\ell 2}, ||\bm \mu_{\ell 1}||_{\infty}\le \alpha_{\ell}, ||\bm \mu_{\ell 2}||_{\infty}\le 1- \alpha_\ell, \forall \ell\in [m]}, \\
		&w_U\bm{x} - (w_U-w_L)\bigg( \sum_{\ell\in [m]} 2^{-\ell}\alpha_{\ell} \bm{x}\bigg)=w_U\bm{x} -(w_U-w_L)\sum_{\ell\in [m]} 2^{-\ell}\bm \mu_{\ell1}.
		\end{align*}
		
		\item Finally, following the approximation and linearization results in Step 3, the equality constraint $\sum_{i\in [n]} \bm c_i \bm c_i^{\top}\bm \sigma_{i1}=w\bm x$ in \eqref{spcacom3_eigen} might not hold exactly.
%
Thus we replace the equality by the following inequality 
		\begin{align*}
		&\bigg|\bigg|\sum_{i \in [n]}\bm{c}_i \bm{c}_i^{\top}\bm\delta_{i1} - w_U \bm{x} + (w_U-w_L)\sum_{i\in [m]} 2^{-i}\bm \mu_{i1}\bigg|\bigg|_{\infty} 
		\\
		&=\bigg|\bigg|\sum_{i \in [n]}\bm{c}_i \bm{c}_i^{\top}z_i\bm x - w_U \bm{x} + (w_U-w_L)\sum_{i\in [m]} 2^{-i}\alpha_i\bm x\bigg|\bigg|_{\infty} \\
		&=\bigg|\bigg|w\bm x - w_U \bm{x} + (w_U-w_L)\sum_{i\in [m]} 2^{-i}\alpha_i\bm x\bigg|\bigg|_{\infty} 
		\leq (w_U-w_L)/2^{m}\leq \epsilon,
		\end{align*}
		which holds for any feasible solution of formulation \eqref{spcacom3_eigen}. 
		
		First, we have $\hat{w}(\epsilon)\geq w^*-\epsilon$ since $w:=w^*-\epsilon$ is feasible to the MILP \eqref{spca_milp}.
		
		Moreover, given an optimal solution $(\hat{\bm{x}},\hat{\bm{z}},\hat{w}(\epsilon))$ to the MILP \eqref{spca_milp}, we must have 
		\begin{align*}
		&\bigg|\bigg|\sum_{i \in [n]}\hat{z}_i\bm{c}_i \bm{c}_i^{\top} \hat{\bm{x}} - \hat{w}(\epsilon)\hat{\bm{x}}\bigg|\bigg|_{\infty} \leq \epsilon\\
		(\Rightarrow)\quad & \min_{\bm{x}:\|\bm{x}\|_\infty=1}\bigg|\bigg|\sum_{i \in [n]}\hat{z}_i\bm{c}_i \bm{c}_i^{\top} \bm{x} - \hat{w}(\epsilon)\bm{x}\bigg|\bigg|_{\infty} \leq \epsilon\\
		(\Rightarrow)\quad & d^{-1/2}\min_{\bm{x}:\|\bm{x}\|_\infty=1}\bigg|\bigg|\sum_{i \in [n]}\hat{z}_i\bm{c}_i \bm{c}_i^{\top} \bm{x} - \hat{w}(\epsilon)\bm{x}\bigg|\bigg|_{2} \leq \epsilon\\
		(\Rightarrow)\quad & d^{-1/2}\min_{\bm{x}:\|\bm{x}\|_2\geq 1}\bigg|\bigg|\sum_{i \in [n]}\hat{z}_i\bm{c}_i \bm{c}_i^{\top} \bm{x} - \hat{w}(\epsilon)\bm{x}\bigg|\bigg|_{2} \leq \epsilon\\
		(\Leftrightarrow)\quad & d^{-1/2}\min_{\bm{x}:\|\bm{x}\|_2=1}\bigg|\bigg|\sum_{i \in [n]}\hat{z}_i\bm{c}_i \bm{c}_i^{\top} \bm{x} - \hat{w}(\epsilon)\bm{x}\bigg|\bigg|_{2} \leq \epsilon
		\end{align*}
		where the first implication is due to $\|\hat{\bm{x}}\|_\infty=1$, the second one is due to $\|\bm{x}\|_{\infty}\geq d^{-1/2}\|\bm{x}\|_{2}$ since $\bm x\in \Re^d$, the third one is because $\|\bm{x}\|_\infty=1$ implies $\|\bm{x}\|_2\geq 1$, and the equivalence is because of monotonicity and positive homogeneity of the objective function. According to the last inequality, there exists an eigenvalue $w$ of matrix $\sum_{i \in [n]}\hat{z}_i\bm{c}_i \bm{c}_i^{\top}$ such that $|\hat{w}(\epsilon)-w|\leq \epsilon\sqrt{d}$, which further implies that $\hat{w}(\epsilon)-w^*\leq \epsilon\sqrt{d}$ since $w\leq w^*$.\qed
	\end{enumerate}
\end{proof}
\subsection{Proof of \Cref{thm_approx_milp}}\label{proof_approx_milp}
\approxmilp*
\begin{proof}From the proof of \Cref{them_milp}, we know that $\overline{w}_5(\epsilon)\leq \overline{w}_5(0)+\epsilon\sqrt{d}$. Thus, it is sufficient to show that
	$$\overline{w}_5(0)\leq k(\sqrt{d}/2+1/2)w^* .$$
	
	We observe that when $\epsilon =0$, the resulting formulation by relaxing binary variables $
	\bm{z}$ to be continuous becomes:
	\begin{align}\label{spcacom3_eigen_eps0}
	\overline{w}_5(0)=\max_{\begin{subarray}{c}
		w, \bm{z} \in \overline{Z}, \bm{x}, \\
		\{\bm \delta_{i1}\}_{i\in [n]},\{\bm \delta_{i2}\}_{i\in [n]}
		\end{subarray}}&\bigg\{w: \sum_{i \in [n]}\bm{c}_i\bm{c}_i^\top \bm{\delta}_{i1}=w\bm{x},\|\bm{x}\|_{\infty}=1,\notag\\
	&{\bm{x} = \bm \delta_{i1}+\bm \delta_{i2}, ||\bm \delta_{i1}||_{\infty}\le z_i, ||\bm \delta_{i2}||_{\infty}\le 1- z_i, \forall i \in [n]}\bigg\},
	\end{align}
	
	Next, we split the proof into three steps.
	\begin{enumerate}[{\bf Step} 1.]
		\item For any feasible solution to problem \eqref{spcacom3_eigen_eps0}, we have
		\begin{align*}
		w&= \frac{||\sum_{i \in [n]}\bm{c}_i \bm{c}_i^{\top}\bm{\delta}_{i1} ||_{\infty}}{||\bm{x}||_{\infty}} = ||\sum_{i \in [n]}\bm{c}_i \bm{c}_i^{\top}\bm{\delta}_{i1} ||_{\infty} \le \sum_{i \in [n]} ||\bm{c}_i \bm{c}_i^{\top}\bm{\delta}_{i1}||_{\infty} = \sum_{i \in [n]} ||\bm{c}_i||_{\infty}|\bm{c}_i^{\top}\bm{\delta}_{i1}| \\
		&\le \sum_{i \in [n]} ||\bm{c}_i||_{\infty}||\bm{c}_i||_{1}||\bm{\delta}_{i1}||_{\infty} \le \sum_{i \in [n]} ||\bm{c}_i||_{\infty}||\bm{c}_i||_{1} z_i \le k\max_{i \in [n]} ||\bm{c}_i||_{\infty}||\bm{c}_i||_{1},
		\end{align*}
		where the first inequality is due to triangle inequality, the second one is because of Holder's inequality, the third one is because $||\bm{\delta}_{i1}||_{\infty}\leq z_i$, and the last one is due to $||\bm{c}_i||_{\infty}||\bm{c}_i||_{1} \le \max_{j \in [n]} ||\bm{c}_j||_{\infty}||\bm{c}_j||_{1}$ for each $i\in [n]$ and $\sum_{i\in [n]}z_i=k$.
		
		\item Now it remains to show that for each $i \in [n]$
		\[ ||\bm{c}_i||_{\infty}||\bm{c}_i||_{1} \le \frac{\sqrt{d}+1}{2} w^*.\]
		
		Let $\varsigma$ be a permutation of index set $[d]$ such that $c_{i,{\varsigma}(1)}, \cdots, c_{i,{\varsigma}(d)}$ are sorted in an ascending order. Then we have
		\begin{align*}
		c_{i,{\varsigma}(1)}^2 + \frac{1}{d-1} \bigg(\sum_{j\in [2,d]}|c_{i,{\varsigma}(j)}|\bigg)^2 \le c_{i,{\varsigma}(1)}^2 + \cdots + c_{i,{\varsigma}(d)}^2 = ||\bm c_i||_2^2 \le w^* , 
		\end{align*}
		where the first inequality is from the arithmetic and quadratic mean inequality and the second inequality follows from $||\bm c_i||_2^2= \lambda_{\max}(\bm c_i\bm c_i^\top) \leq w^*$.
		
		For ease of exposition, let us introduce $v_1 = |c_{i,{\varsigma}(1)}|$ and $v_2 =\sum_{j\in [2,d]}|c_{i,{\varsigma}(j)}|$.
		Next, let us consider an optimization problem
		\begin{equation}\label{eq_parameter_milp}
		\nu=\max_{\bm{v}\in \Re_+^2}\bigg\{ v_1(v_1+v_2): v_1^2 + 1/(d-1) v_2^2\le w^*\bigg\},
		\end{equation}
		whose optimal value clearly provides an upper bound of $||\bm{c}_i||_{\infty}||\bm{c}_i||_{1}$.
		
		To solve \eqref{eq_parameter_milp}, we first rewrite $v_1,v_2$ as
		\begin{align*}
		v_1 = r\sin (\theta) r , v_2 = r\sqrt{d-1}\cos(\theta) , \theta \in [0,\pi/2], r\leq \sqrt{w^*}.
		\end{align*}
		In this way, the objective function \eqref{eq_parameter_milp} is equal to
		\begin{align*}
		v_1(v_1 + v_2) &= v_1^2 + v_1v_2 = r^2 \sin^2(\theta) + r^2\sqrt{d-1} \sin(\theta) \cos(\theta) = r^2\frac{1-\cos(2\theta)}{2} + r^2 \sqrt{d-1} \frac{\sin(2\theta)}{2}\\
		& =\frac{r^2}{2}-\frac{r^2}{2}\cos(2\theta)+\frac{1}{2}r^2 \sqrt{d-1} \sin(2\theta)\le \frac{1}{2}r^2 + \frac{\sqrt{d}}{2}r^2 \le \frac{\sqrt{d}+1}{2} w^*,
		\end{align*}
		where the first inequality is due to Cauchy-Schwartz inequality and the second one is because $r^2\leq w^*$. Thus, we must have
		$$||\bm{c}_i||_{\infty}||\bm{c}_i||_{1}\leq \frac{\sqrt{d}+1}{2} w^*.$$
				This proves the first bound $k (\sqrt{d}/2+1/2)$ together with Step 1.
		\item 	We now prove the second bound.
	Plugging the equations $\bm{\delta}_{i1} =\bm x- \bm{\delta}_{i2}$ for all $i \in [n]$, we rewrite the continuous relaxation value as
		\begin{align*}
		w&= \frac{||\sum_{i \in [n]}\bm{c}_i \bm{c}_i^{\top}(\bm x - \bm{\delta}_{i2}) ||_{\infty}}{||\bm{x}||_{\infty}} \le \frac{||\sum_{i \in [n]}\bm{c}_i \bm{c}_i^{\top}\bm x ||_{\infty}}{||\bm{x}||_{\infty}} + \frac{||\sum_{i \in [n]}\bm{c}_i \bm{c}_i^{\top} \bm{\delta}_{i2} ||_{\infty}}{||\bm{x}||_{\infty}}
		\\
		&\le \frac{||\sum_{i \in [n]}\bm{c}_i \bm{c}_i^{\top}\bm x ||_{\infty}}{||\bm{x}||_{\infty}} 
		+(n-k)\frac{\sqrt{d}+1}{2} w^* \le \max_{i\in [d]}\sum_{j \in [d]} | \overline{C}_{ij}| + (n-k)\frac{\sqrt{d}+1}{2} w^* ,
		\end{align*}
		where $\overline{\bm C} := \bm C \bm C^{\top} = \sum_{i\in [n]}\bm c_i \bm c_i^{\top}$ and the first inequality is from the triangle inequality, the second one follows from the derivations in Steps 1 and 2, and the third one is due to $x_i \le 1$ for each $i\in [d]$.
		
		Next, the first term of the right-hand side above can be upper bounded by 
		\begin{align*}
		 \max_{i\in [d]}\sum_{j \in [d]} | \overline{C}_{ij}| = ||\overline{\bm C}||_1	\le \sqrt{d} ||\overline{\bm C}||_2 =\sqrt{d} \lambda_{\max}(\overline{\bm C}) 
		\le \frac{n}{k} \sqrt{d} w^* ,
		\end{align*}
	where the equations are from the definition of $\ell_1$-norm and $\ell_2$-norm of a matrix and the second inequality is due to $\lambda_{\max}(\overline{\bm C})=\lambda_{\max}(\bm A) \le n/k w^* $.	\qed

	\end{enumerate}
\end{proof}

\subsection{Proof of \Cref{lem:r1ssvd}} \label{proof_lem_r1ssvd}
\lemrssvd*
\begin{proof} 
The proof includes two parts.
\begin{enumerate}[(i)]
\item By the definition of augmented matrix $\overline{ \bm{A} }$ in \eqref{eqmatrixA}, for its submatrix $ \overline{ \bm{A} }_{S, S}$, we observe that
	\[
	\overline{\bm{A}}_{S,S}=
	\begin{bmatrix} 
	\bm 0 & \bm A_{S_1, S_2} \\
	\bm{A}^{\top}_{S_1, S_2} &\bm 0 
	\end{bmatrix}
.\]
Then the statement in Part (i) directly follows from the result in \citet{ben2001lectures}, which shows that the eigenvalues of an augmented symmetric matrix exactly are equal to the singular values and negative ones of the original matrix.
 \item 	The first equality $\lambda_{\max}(\overline{\bm{A}}_{S,S})=\sigma_{\max}(\bm A_{S_1,S_2})$ is obtained from Part (i).
 
For the largest singular value of $\bm A_{S_1,S_2}$, we have
 \begin{align} \label{eq_singular}
 \sigma_{\max}(\bm A_{S_1,S_2}) &= \max_{\bm u \in \Re^{k_1}, \bm v\in \Re^{k_2}} \left\{\bm u^{\top} \bm A_{S_1,S_2} \bm v: ||\bm u ||_2=1, ||\bm u||_2 =1 \right\} \notag \\
 &= \frac{1}{2}\max_{\bm{x}\in \Re^{k_1+k_2}} \left\{ \bm{x}^{\top}
 \overline{\bm A}_{S,S}
 \bm{x} : ||\bm{x}_{1:k_1}||_{2} =1, ||\bm{x}_{k_1+1:k_1+k_2}||_{2}=1\right\},
 \end{align}
 which proves the second equality of Part (ii).
 
As for the last equality of Part (ii), we let $\hat{w}^*_{\rm SVD}$ denote the optimal value of the right-hand side SDP problem. 
%
 Then we must have $\hat{w}^*_{\rm SVD}\geq \sigma_{\max}(\bm A_{S_1,S_2})$ as the SDP problem is exactly a SDP relaxation of the maximization problem over $\bm x$ in \eqref{eq_singular} by relaxing the rank-one constraint.
 On the other hand, summing up two constraints in the SDP problem, we obtain an upper bound of $\hat{w}^*_{\rm SVD}$, i.e.,
 \[
 \hat{w}^*_{\rm SVD}\leq \frac12\max_{\bm{X} \in \S_+^{k_1+k_2}} \left \{ \tr(\overline{\bm{A}}_{S,S}\bm{X}) : \tr(\bm{X})=2 \right \}=\lambda_{\max}(\overline{\bm{A}}_{S,S})=\sigma_{\max}(\bm A_{S_1,S_2}),\]
 where the first equality is due to Part (ii) in \Cref{lemspca}.\qed
\end{enumerate}
\end{proof}

\subsection{Proof of \Cref{thm_model2_relax_svd}} \label{proof_thm_model2_relax_svd}
\themonessvd*
\begin{proof}
For the matrix $\overline{ \bm{A} }^{\#}$ defined in \eqref{eq_Astar}, using Part (i) in \Cref{lem:r1ssvd}, we can derive that its largest eigenvalue is equal to 2$\sigma_{\max}(\bm A)$. Let $(\hat{\bm z}, \hat{\bm X}, \hat{\bm W}_1, \cdots, \hat{\bm W}_{m+n} )$ denote an optimal solution to the continuous SDP relaxation of problem \eqref{sdp_two_eq_svd}. We now have
	\begin{align*}
	2\sigma_{\max}(\bm A) = \lambda_{\max}\left(\overline{ \bm{A} }^{\#}\right)=\max_{\bm X \succeq 0, \tr(\bm X)=1} \bigg\{ \sum_{i\in [m+n]} \bm c_i^{\top} \bm X \bm c_i \bigg\} \ge \sum_{i\in [m+n]} \bm c_i^{\top} \hat{\bm X} \bm c_i \ge \sum_{i\in [m+n]} \bm c_i^{\top} \hat{\bm W_i} \bm c_i,
	\end{align*}
	where the last inequality is because $ \hat{\bm X} \succeq \hat{\bm W}_i$ for each $i \in [m+n]$. Note that the right-hand side above is equal to $\overline{w}_{\rm SVD1} + \sigma_{\max}(\bm A)$ and the inequalities above lead to
	\begin{align*}
	\overline{w}_{\rm SVD1} = \sum_{i\in [m+n]} \bm{c}_i^{\top} \hat{\bm W_i} \bm{c}_i -\sigma_{\max} (\bm A) \le 2\sigma_{\max} (\bm A) - \sigma_{\max} (\bm A)= \sigma_{\max} (\bm A). 
	\end{align*}
	Now it remains to show that
\begin{claim}\label{claim:singular}
	$\sigma_{\max} (\bm A) \le \sqrt{mnk_1^{-1}k_2^{-1}}  w^*_{\rm SVD}$.
\end{claim}
\begin{proof}
	Let $\bm{u}_1$, $\bm{v}_1$ denote the top right and left eigenvectors of $\bm{A}$, i.e., $\bm{u}_1^{\top} \bm{A} \bm{v}_1 =\sigma_{\max}(\bm A), \bm{A} \bm{v}_1 = \sigma_{\max}(\bm{A})\bm{v}_1,\bm{u}_1^{\top} \bm{A} = \sigma_{\max}(\bm{A})\bm{u}_1^{\top} $. We tailor $\bm{u}_1$, $\bm{v}_1$ to meet the feasibility of R1-SSVD \eqref{ssvd} as below
	\begin{align*}
	&\hat{u}_{j1}=
	\begin{cases}
{u}_{j1}, \ \ \ \ &\text{\rm if $ {u}_{j1}$ is one of the $k_1$ largest entries of $\bm{u}_1$} \\
	0, \ \ \ \ &\text{\rm otherwise}
	\end{cases},\forall j\in [n],\\
	&\hat{v}_{j1}=
	\begin{cases}
	(\bm{A}^\top \hat{\bm u})_j, \ \ \ \ &\text{\rm if $|(\bm{A}^\top \hat{\bm u})_j|$ is one of the $k_2$ largest entries of $|\bm{A}^\top \hat{\bm u}|$} \\
	0, \ \ \ \ &\text{\rm otherwise}
	\end{cases},\forall j\in [m].
	\end{align*}

	Let us normalize $\hat{\bm{u}}_1= \frac{\hat{\bm{u}}_1}{||\hat{\bm{u}}_1||_2}$ and $\hat{\bm{v}}_1= \frac{\hat{\bm{v}}_1}{||\hat{\bm{v}}_1||_2}$. Clearly, $(\hat{\bm{u}}_1,\hat{\bm{v}}_1)$ is feasible R1-SSVD \eqref{ssvd}. Then we have
	\begin{align*}
\sqrt{\frac{k_1}{n}}\sigma_{\max}(\bm{A}) \le 	\sigma_{\max}(\bm{A}) \hat{\bm{u}}_1^{\top} {\bm{u}}_1 = \hat{\bm{u}}_1^{\top} \bm{A} {\bm{v}}_1 \le \|\hat{\bm{u}}_1^{\top} \bm{A}\|_2 \le \sqrt{\frac{m}{k_2}} \hat{\bm{u}}_1^{\top} \bm{A} \hat{\bm{v}}_1 \le \sqrt{\frac{m}{k_2}} w^*_{\rm SVD},
	\end{align*}
	where the first inequality is due to the definition of $\hat{\bm{u}}_1 $, the equality is because of the definition of ${\bm{v}}_1$, the second inequality is due to the Cauchy-Schwartz inequality, the third one is based on the choice of $\hat{\bm v}_1$, and the last one is due to the feasibility of $(\hat{\bm{u}}_1,\hat{\bm{v}}_1)$.
	This completes the proof.
%
%
\qedA
\end{proof}
\qed
\end{proof}

\subsection{Proof of \Cref{svdlem_ineq}} \label{proof_svdlem_ineq}
\svdlemineq*

\begin{proof} 
According to \Cref{svdthem_eq}, there must exist an optimal solution $(\bm{z}^*,\bm X^*)$ to MISDP \eqref{svd_sdp_one} such that $\bm{X}^*$ is rank-one. Thus, without loss of generality, for any feasible solution $(\bm{z},\bm X)$ of SPCA \eqref{sdp_one}, we can assume that $\bm X = \begin{bmatrix} 
\bm{u}\\
\bm{v}
\end{bmatrix}\begin{bmatrix} 
\bm{u}\\
\bm{v}
\end{bmatrix}^{\top}$, where vectors $( \bm{u},\bm{v})$ thus satisfy 
\begin{align*}
||\bm u||_2 = ||\bm v||_2 =1, ||\bm u||_1 \le \sqrt{k_1}, ||\bm v||_1 \le \sqrt{k_2}.
\end{align*}
Then the rest of the proof is almost identical to that of \Cref{lem_ineq} and is thus omitted for brevity.
%
%
%
\end{proof}

\subsection{Proof of \Cref{thm_truncation}} 
\label{proof_thm_truncation}
\thmtruncation*
\begin{proof}
	We derive the three approximation ratios of the truncation algorithm below.
\begin{enumerate}[(i)]
\item 

According to the truncation in the standard basis, the obtained vector $\hat{\bm{u}}_i$ is feasible to the R1-SSVD problem for each $i \in [n]$ and is also optimal to the following problem 
\begin{align*}
\hat{\bm{u}}_i\in \arg\max_{||\bm{u}_i||_2=1,||\bm{u}_i||_0=k_1} \left \{ \bm{u}_i^{\top} \bm{A} \bm{e}_i\right \}, \forall i \in [n].
\end{align*}

	Suppose the optimal solution of the R1-SSVD \eqref{ssvd} to be $\bm{u}^*$ and $\bm{v}^*$, let $S_1^*, S_2^*$ denote their supports, respectively. We then rewrite $\bm{v}^*=\sum_{i\in S_2^*} v^*_i \bm{e}_i$ and we have
\begin{align*}
w_{\rm SVD}^* = (\bm{u}^*)^{\top}\bm{A}\bm{v}^* = \sum_{i\in S_2^*} v^*_i (\bm{u}^*)^{\top} \bm{A} \bm{e}_i \le \sqrt{\sum_{i\in S_2^*} (v^*_i)^2} \sqrt{\sum_{i\in S_2^*} [(\bm{u}^*)^{\top} \bm{A} \bm{e}_i]^2} \le \sqrt{k_2} \max_{i\in [n]}\hat{\bm{u}}_i^{\top} \bm{A} \bm{e}_i,
\end{align*}
where the first inequality is due to Cauchy-Schwartz and the second one is because of maximality of $\max_{i\in [n]}\hat{\bm{u}}_i^{\top} \bm{A} \bm{e}_i$.

Since $(1-(k_2-1)\epsilon)\bm{e}_i+\epsilon \sum_{j\in [k_2]\cup\{i\}\setminus\{i\}}\e_j$ with sufficiently small $\epsilon>0$ is feasible to R1-SSVD \eqref{ssvd}, thus the right-hand side above is an lower bound of R1-SSVD according to the continuity by letting $\epsilon\rightarrow 0$. This prove the approximation ratio $\sqrt{k_2^{-1}}$.

Similarly, we can derive
\begin{align*}
w_{\rm SVD}^* = (\bm{u}^*)^{\top}\bm{A}\bm{v}^* \le \sqrt{k_1} \max_{j\in [m]}{\bm{e}}_j^{\top} \bm{A} \hat{\bm{v}}_j,
\end{align*}
which prove the approximation ratio $\sqrt{k_1^{-1}}$.
\item Following the proof of \Cref{claim:singular}, for the truncation in the eigen-space basis, we have
	\begin{align*}
\sqrt{\frac{k_1}{n}}w^*_{\rm SVD} \le \sqrt{\frac{k_1}{n}}\sigma_{\max}(\bm{A}) \le 	\sigma_{\max}(\bm{A}) \hat{\bm{u}}_1^{\top} {\bm{u}}_1 = \hat{\bm{u}}_1^{\top} \bm{A} {\bm{v}}_1 \le \|\hat{\bm{u}}_1^{\top} \bm{A}\|_2 \le \sqrt{\frac{m}{k_2}} \hat{\bm{u}}_1^{\top} \bm{A} \hat{\bm{v}}_1,
\end{align*}
which proves the approximation ratio of $\sqrt{k_1k_2m^{-1}n^{-1}}$.\qed
\end{enumerate}

\subsection{Proof of \Cref{thm_svdgreedy}} \label{proof_thm_svdgreedy}
\thmsvdalg*
\begin{proof}
The proof is split into two parts.
\begin{enumerate}[(i)]
    \item In R1-SSVD \eqref{eqssvdcom}, according to the part (i) of the proof of \Cref{thm_truncation}, we have
	\[w^*_{\rm SVD}  \le \sqrt{k_2} \max_{j\in [n]} \hat{\bm u}_j^{\top} \bm A \bm e_j \le \sqrt{k_1k_2} \max_{i\in [m],j\in [n]}\bm A_{ij} ,\] 	
	where vectors $\{\hat{\bm u}_i  \}_{i\in [n]}\subseteq \Re^m$ are obtained by the normalized $k_1$-truncation in the standard basis of $\bm A$. Then, following the similar analyses of \Cref{prop_greedy_alg}
and \Cref{prop_LS_alg}, the largest singular value from greedy \Cref{alg:svd_greedy} and local search \Cref{alg:svd_localsearch} must be lower bounded by  $\max_{i\in [m],j\in [n]}\bm A_{ij} $.
\item We next show an example in which the ratio $\sqrt{k_1^{-1}k_2^{-1}}$ can be achieved. Suppose that, without loss of generality, $k_1\leq k_2$. Then, consider $m=2k_2$, $n=2k_2$, and  matrix $\bm A \in \Re^{m\times n}$ as
\begin{align*}
\bm A:=\begin{bmatrix} 
&\bm{I}_{k_2} & \bm{0}_{k_2 \times k_2}\\
&\bm{0}_{k_2 \times k_2} & \bm{1}_{k_2 \times k_2}\\
\end{bmatrix}.
\end{align*}
Above, the submatrix $\bm{A}_{[k_1], [k_2]}$ satisfies greedy and local optimality conditions with the objective value equal to 1, while the best size $k_1\times k_2$ submatrix is $\bm{A}_{[k_2+1,k_2+k_1], [k_2+1, 2k_2]}$ with the optimal value $\sqrt{k_1k_2}$. \qed
\end{enumerate}

\end{proof}

\end{proof}

\end{appendices}
\end{document}